\documentclass[lettersize,journal]{IEEEtran}
\usepackage{url}            
\usepackage{booktabs}       
\usepackage{amsfonts}       
\usepackage{nicefrac}       
\usepackage{microtype}      
\usepackage{xcolor}         
\usepackage{amsmath}
\usepackage{graphicx}
\usepackage{algorithm}
\usepackage{algorithmic}
\usepackage{multicol}
\usepackage{multirow}
\usepackage{stfloats}
\makeatletter

\newcommand{\Rmnum}[1]{\expandafter\@slowromancap\romannumeral #1@}
\makeatother
\newcommand{\etal}{{\emph{et al.}}}
\newcommand{\ie}{{\emph{i.e.}}}
\newcommand{\eg}{{\emph{e.g.}}}

\usepackage{bm}

\usepackage[pagebackref=false,colorlinks=true, linkcolor = red,citecolor=blue]{hyperref}
\usepackage{tikz}
\definecolor{lime}{HTML}{A6CE39}
\DeclareRobustCommand{\orcidicon}{%
    \begin{tikzpicture}
    \draw[lime, fill=lime] (0,0) 
    circle [radius=0.16] 
    node[white] {{\fontfamily{qag}\selectfont \tiny ID}};    \draw[white, fill=white] (-0.0625,0.095) 
    circle [radius=0.007];    \end{tikzpicture}
    \hspace{-2mm}}
\foreach \x in {A, ..., Z}{%
    \expandafter\xdef\csname orcid\x\endcsname{\noexpand\href{https://orcid.org/\csname orcidauthor\x\endcsname}{\noexpand\orcidicon}}
}

\begin{document}

\title{Asynchronous Multimodal Video Sequence Fusion via Learning Modality-Exclusive and -Agnostic Representations}


\author{Dingkang Yang\orcidA{}, Mingcheng Li\orcidD{}, Linhao Qu, Kun Yang\orcidH{}, Peng Zhai, Song Wang,~\IEEEmembership{Senior Member, IEEE}, \\ Lihua Zhang\orcidC{},~\IEEEmembership{Member, IEEE}

\thanks{This work was supported in part by National Key R\&D Program of China 2021ZD0113503 and in part by Shanghai Municipal Science and Technology Major Project 2021SHZDZX0103. The first three authors contributed equally. (\emph{Corresponding authors: Peng Zhai and Lihua Zhang})} 
\thanks{Dingkang Yang, Mingcheng Li, Kun Yang, Peng Zhai, and Lihua Zhang are with the Academy for Engineering and Technology, Fudan University, Shanghai 200433, China. E-mail: \{dkyang20, kunyang20, pzhai, lihuazhang\}@fudan.edu.cn, mingchengli21@m.fudan.edu.cn.}
\thanks{Linhao Qu is with the Digital Medical Research Center, School of Basic Medical Science, Fudan University, Shanghai 200032, China. E-mail: lhqu20@fudan.edu.cn.}
\thanks{Song Wang is with the Department of Computer Science and Engineering, University of South Carolina, Columbia, SC 29208 USA. E-mail: songwang@cec.sc.edu.}
}

\markboth{IEEE TRANSACTIONS ON CIRCUITS AND SYSTEMS FOR VIDEO TECHNOLOGY}%
{Shell \MakeLowercase{\textit{et al.}}: A Sample Article Using IEEEtran.cls for IEEE Journals}


\maketitle

\begin{abstract}
Understanding human intentions (\eg, emotions) from videos has received considerable attention recently. 
Video streams generally constitute a blend of temporal data stemming from distinct modalities, including natural language, facial expressions, and auditory clues.
Despite the impressive advancements of previous works via attention-based paradigms, the inherent temporal asynchrony and modality heterogeneity challenges remain in multimodal sequence fusion, causing adverse performance bottlenecks.
To tackle these issues, we propose a Multimodal fusion approach for learning modality-Exclusive and modality-Agnostic representations (MEA) to refine multimodal features and leverage the complementarity across distinct modalities. 
On the one hand, MEA introduces a predictive self-attention module to capture reliable context dynamics within modalities and reinforce unique features over the modality-exclusive spaces.
On the other hand, a hierarchical cross-modal attention module is designed to explore valuable element correlations among modalities over the modality-agnostic space.
Meanwhile, a double-discriminator strategy is presented to ensure the production of distinct representations in an adversarial manner.
Eventually, we propose a decoupled graph fusion mechanism to enhance knowledge exchange across heterogeneous modalities and learn robust multimodal representations for downstream tasks.
Numerous experiments are implemented on three multimodal datasets with asynchronous sequences. Systematic analyses show the necessity of our approach.
\end{abstract}


\section{Introduction}
\IEEEPARstart{W}{ith} the popularity of multimedia websites such as YouTube and Vimeo, humans are sharing their opinions and reviews through multimedia videos every day. 
Video streams conventionally consist of time-series data drawn from diverse modalities, embracing attributes like natural language~\cite{zhang2021multimodal}, visual representations~\cite{marivani2022designing}, and acoustic expressions~\cite{zhou2019understanding}.
Researchers have recently focused on analyzing videos from a multimodal perspective to promote a holistic understanding of human intentions and expressions~\cite{feng2022temporal,shen2021bbas}.
The core component for fully exploiting the critical information provided by multiple modalities is to perform the pragmatic fusion of multimodal sequence data.
To this end, previous prominent efforts~\cite{yang2022disentangled,yang2022learning,lei2023text,zadeh2016multimodal,yang2023target,yang2022contextual} have provided promising solutions for multimodal fusion through well-designed strategies and sophisticated structures.
Despite recent impressive advancements, challenges remain due to unpreventable temporal asynchrony and modality heterogeneity.

\textbf{Temporal Asynchrony.}
In practical applications, the collected multimodal streams are usually asynchronous and without alignment, attributed to the variable frequencies at which sequences from different modalities are received.
For example, a video frame depicting a rigid facial expression could potentially correspond to a preceding negative vocalization. Furthermore, the spoken dialogue or the accompanying subtitles might not perfectly align with the visual content.
The inherent temporal asynchrony across modalities leads to severe performance bottlenecks~\cite{tsai2019multimodal}.
Consequently, most previous methods \cite{zadeh2016multimodal,wang2019words,pham2019found,tsai2018learning,wu2021text,liang2018multimodal,rahman2020integrating,wu2023denoising} tackled the above issue via word-level alignment. 
The manual pre-processing is performed on the visual and audio sequences to synchronize the resolution of textual words.
Unfortunately, the alignment process usually involves enormous time and labour overheads, requiring domain-related knowledge engineering.
Moreover, such word-aligned multimodal fusion ignores vital perception clues in the long-range contextual contingencies among modalities.
Recent attention-driven efforts~\cite{tsai2019multimodal,wang2023cross,sahay2020low,lv2021progressive,liang2021attention,yang2023target} can deal directly with asynchronous multimodal sequences benefiting from the advantages of Transformer-like structures~\cite{vaswani2017attention} in modelling the temporal element correlations.
However, they either only considered shallow cross-modal adaptions~\cite{tsai2019multimodal,wang2023cross,sahay2020low,lv2021progressive} or focused on coarse-grained finite interactions~\cite{liang2021attention,yang2023target,lei2023text}.
These inadequate designs ignore the intra-modal dynamics and potentially cause the learned element dependencies to be unreliable.

\textbf{Modality Heterogeneity.} Another dilemma of multimodal sequence fusion is the intrinsic modality heterogeneity~\cite{yang2022disentangled}. 
Different modalities typically utilize varying expressiveness to convey semantic information. For instance, the language modality transcribed from videos has more abstract and informative clues than nonverbal visual signals and acoustic behaviours. As a result, the inconsistent knowledge intensity among strong and weak modalities increases the difficulty of fusing multimodal sequences~\cite{lei2023text}.
Meanwhile, heterogeneous characteristics across different modalities usually introduce information redundancy and distribution gaps, leading to task-irrelevant semantics and fragile multimodal representations.
Several implementations~\cite{Hazarika2020mm,wu2021text,li2023decoupled} have recently disentangled hybrid multimodal features by learning discriminative semantics in distinct modality spaces. 
Nonetheless, these efforts are premature due to rudimentary constraints~\cite{Hazarika2020mm} or non-generalizable strategies~\cite{wu2021text,li2023decoupled} that fail to appreciably capture the modality heterogeneity and remove its interference.

To mitigate the above challenges, we propose an asynchronous Multimodal fusion approach for learning modality-Exclusive and -Agnostic representations (MEA) to refine multimodal features holistically.
The novelty of MEA is reflected in three aspects.
(i) We devise a feature decoupling pattern to comprehend distinct aspects of multimodal representations through the exploration of both modality-exclusive and -agnostic domains.
For the modality-exclusive representations, we introduce a predictive self-attention module to effectively enhance the unique features of each modality and learn the long-range context dynamics within the modality.
For the modality-agnostic representations, we present a hierarchical cross-modal attention module that aims to foster robust inter-modal interactions and capture consequential correlations among constituent elements spanning diverse modalities.
(ii) We present a novel double-discriminator adversarial strategy to supervise the representation production and the corresponding parameter learning.
This strategy suitably ensures that our approach can capture the specific properties of each modality while learning informational commonalities across modalities.
MEA overcomes the \textit{temporal asynchrony} dilemma by capturing intra- and inter-modal element dependencies in exclusive and agnostic subspaces with the above-tailored components.
The systematic analysis confirms the rationality of our components.
(iii) Ultimately, we devise a decoupled graph fusion mechanism to aggregate the complementary strengths among the disentangled representations for mitigating the \textit{modality heterogeneity} dilemma.

MEA is superior to previous state-of-the-art (SOTA) works on several multimodal understanding datasets with asynchronous attributes. Comprehensive experiments show the merit of our approach in the unaligned setting.
The proposed techniques can be readily extended to diverse video computing applications to improve the accuracy of video content understanding. On the one hand, our attention-driven modules can efficiently capture element correlations and context dependencies among different frames on video sequences to enhance the recognition of temporal events (\textit{e.g.}, anomaly video detection).
On the other hand, our approach allows researchers to learn complementary semantics in heterogeneous modalities from the multimodal decoupling perspective, leading to pragmatic and informative video representations.

This work significantly extends our preliminary conference version~\cite{yang2022learning}. We offer multi-faceted improvements to strengthen our work further and promote more far-reaching impacts.
Concretely, (i)) we introduce the Hilbert-Schmidt Independence Criterion constraint to encourage exclusive and agnostic encoders to learn different aspects of multimodal data better. The new disparity constraint has more robust convergence and separation of different representations compared to the previous regularization; (ii) we devise a new decoupled graph fusion mechanism to facilitate knowledge exchange and information aggregation between heterogeneous and homogeneous multimodal representations. Our novel fusion mechanism further bridges the performance bottleneck caused by the modality heterogeneity; (iii) we implement additional sensitivity analyses to investigate the robustness of MEA against different factors and parameters; (iv) we compare more and newer SOTA models to comprehensively assess the broad effectiveness of the proposed approach on different dataset benchmarks; 
(v) the current version provides a more in-depth discussion and introduction regarding our motivation, network structure, and implementation; 
(vi) more comparative and qualitative experiments are conducted to evaluate our model. 
Last but not least, we empirically observe that our MEA outperforms the recent algorithms in the challenging unaligned setting.

The rest of this paper is organized as follows. In Section~\ref{sec2}, we discuss related works in terms of multimodal sequence fusion and multimodal representation learning. Section~\ref{sec3} describes in detail the different modules, mechanisms, and strategies for tight collaboration in the proposed methodology in a general-division structure. In addition, the optimization objective is clarified. The dataset configuration for evaluation and implementation details for model training are provided in Section~\ref{sec4}. In Section~\ref{sec5}, we give extensive experimental results and corresponding discussions.
Finally, our conclusion and limitation are drawn in Section~\ref{sec6}.

\section{Related Work}
\label{sec2}
\subsection{Multimodal Video Computing}
Video computing~\cite{liu2023stochastic,liu2024memory} aims at extracting task-relevant feature semantics from video clips or intercepted images in a learning-based manner to serve diverse downstream tasks, such as dynamic facial expression recognition~\cite{zhang2023transformer}, extreme video summarisation~\cite{tang2023tldw}, and driver intention monitoring~\cite{yang2023aide}. 
Benefiting from rich multimedia resources, researchers have gradually utilized additional tools to extract multimodal signals~\cite{he2023multimodal,fu2022drake,guan2023egocentric,li2023towards,li2024unified,yang2024SuCi,liu2024generalized,MRG3Net10419057,yang2024TSIF}, including audio and transcribed texts from videos, to tackle the perception dilemmas when the video modality is blurred and occluded.
He~\etal~\cite{he2023multimodal} presented a multimodal mutual attention-based sentiment analysis framework to explore the unique and public semantics across multiple modalities from sophisticated video contexts.
This framework includes a multi-view hierarchical fusion module to accomplish full multimodal data fusion under conditions where different modalities are constrained to each other. Subsequently, the fusion order is adjusted to enhance cross-modal complementarity.
Tang~\etal~\cite{tang2023tldw} introduced an unsupervised hierarchical optimal transport network to extract extreme video summarization from predefined video-document pairs, which consists of hierarchical multimodal encoders, hierarchical
multimodal fusion decoders, and optimal transport
solvers.
In addition, Guan~\etal~\cite{guan2023egocentric} devised a multimodal transformer-driven dual-action prediction model to predict near-future actions from videos captured from a first-person perspective. A two-stage training scheme is simultaneously considered to enhance the correlation and consistency between observed and unobserved video clips.

\subsection{Multimodal Sequence Fusion}

Video understanding as a multimedia medium for linking human-computer interaction applications requires the fusion of time-series data from multiple modalities,
such as natural language, facial gestures, and acoustic behaviors.
A predominant focus in earlier studies~\cite{huang2021emotion,duan2021multi} is the multimodal fusion of stationary attributes extracted from video clips, frequently disregarding the underlying relationships that interlink elements across multimodal sequences.
Nevertheless, multimodal flows usually have an asynchronous dilemma since distinct sensors in real-world applications have variable sampling rates for different modality sequences.
To tackle this issue, some remarkable efforts~\cite{zadeh2016multimodal,yang2022disentangled,yang2022contextual,wang2019words,pham2019found,tsai2018learning,wu2021text,liang2018multimodal,rahman2020integrating,wu2023denoising,yang2023aide,li2024unified,li2023towards,yang2024SuCi,yang2024MCIS} have attempted to align visual and audio sequences with textual words via manual pre-processing before multimodal fusion.
These representative works include shared-private feature learning~\cite{wu2021text}, cyclic translation mechanism~\cite{pham2019found}, denoising bottleneck structure~\cite{wu2023denoising}, recurrent multistage fusion~\cite{liang2018multimodal}, nonverbal temporal interaction~\cite{wang2019words}, etc. 
For example, Wu~\etal~\cite{wu2023denoising} designed a video multimodal fusion method by the enoising bottleneck with mutual information maximization.
Unfortunately, manual alignment is catastrophic for industrial applications due to the high amount of labour and time required. The aligned procedure also ignores context-aware clues.

Recently, several works~\cite{tsai2019multimodal,wang2023cross,sahay2020low,lv2021progressive,liang2021attention,yang2024towards} attempted to deal directly with asynchronous multimodal sequences by the attention-driven paradigm.
Compared to the recurrent
neural network~\cite{medsker2001recurrent} and long short-term memory network~\cite{hochreiter1997long} based structures, the Transformer-like architecture is more effective in exploring the element correlations. 
For instance,
Tsai~\textit{et al.}~\cite{tsai2019multimodal} proposed the Multimodal Transformer (MulT) to achieve directed temporal adaptation and cross-modal interactions between two modalities. Then,
Sahay~\etal~\cite{sahay2020low} improved the computational efficiency and enriched the fusion granularity of MulT by introducing low-rank matrix factorization~\cite{liu2018efficient}.
Lv~\etal~\cite{lv2021progressive} introduced a message hub to explore three-way multimodal fusion and enhance low-level features of source modalities.
However, simple attention is insufficient due to information redundancy and distribution gap across modalities, which may cause the learned correlations to be unreliable.
In contrast, the proposed approach aims to capture intra-modal dynamics and cross-modal interactions in modality-decoupled spaces through two tailored attention modules, bridging the shortcomings of previous works.

\subsection{Multimodal Representation Learning}

Unlike the isolated modality~\cite{yang2024pediatricsgpt,yang2024robust,yang2023how2comm,du2021learning,yang2023context,liu2024generalized,wang2022spacenet},
the heterogeneity among different modalities tends to increase the difficulty of analyzing the same expression in multimodal human languages~\cite{yang2022disentangled}.
For this purpose, learning how to extract informative multimodal representations has attracted widespread attention in recent years~\cite{zhang2019multimodal,yu2021learning,zhang2022tailor,li2023decoupled, park2016image,2020Learning,bousmalis2016domain,chen2024can,chen2024efficiency,chen2024detecting,jiang2024medthink}.
For example, Gwangbeen~\etal~\cite{park2016image} applied the adversarial concept to multimodal learning and implemented multimodal embeddings through the category information. 
Yu~\etal~\cite{yu2021learning} captured modality-specific representations through auxiliary self-supervised multi-task learning. They also introduced a weight accommodation strategy to balance the learning procedure across different subtasks.
To improve multimodal representations, Sun~\etal~\cite{2020Learning} utilized the deep exemplary correlation analysis to establish high-level connections between text-based audio and text-based video.
Furthermore, Bousmalis~\etal~\cite{bousmalis2016domain} designed a domain separation network to extract distinct representations by explicitly modelling the shared and domain-specific private features of source and target domains. 
More recently, Li~\etal~\cite{li2023decoupled} introduced a multimodal distillation paradigm to achieve cross-modal knowledge transfer and exchange.
In comparison, the proposed MEA learns modality-exclusive and modality-agnostic multimodal representations by a double-discriminator adversarial strategy.
These different features reveal the diversity and commonality among heterogeneous modalities, facilitating learning more pragmatic and robust multimodal representations from a complementary perspective.

\section{Methodology}
\label{sec3}
\subsection{Model Overview}
\begin{figure*}[t]
  \centering
  \includegraphics[width=0.75\linewidth]{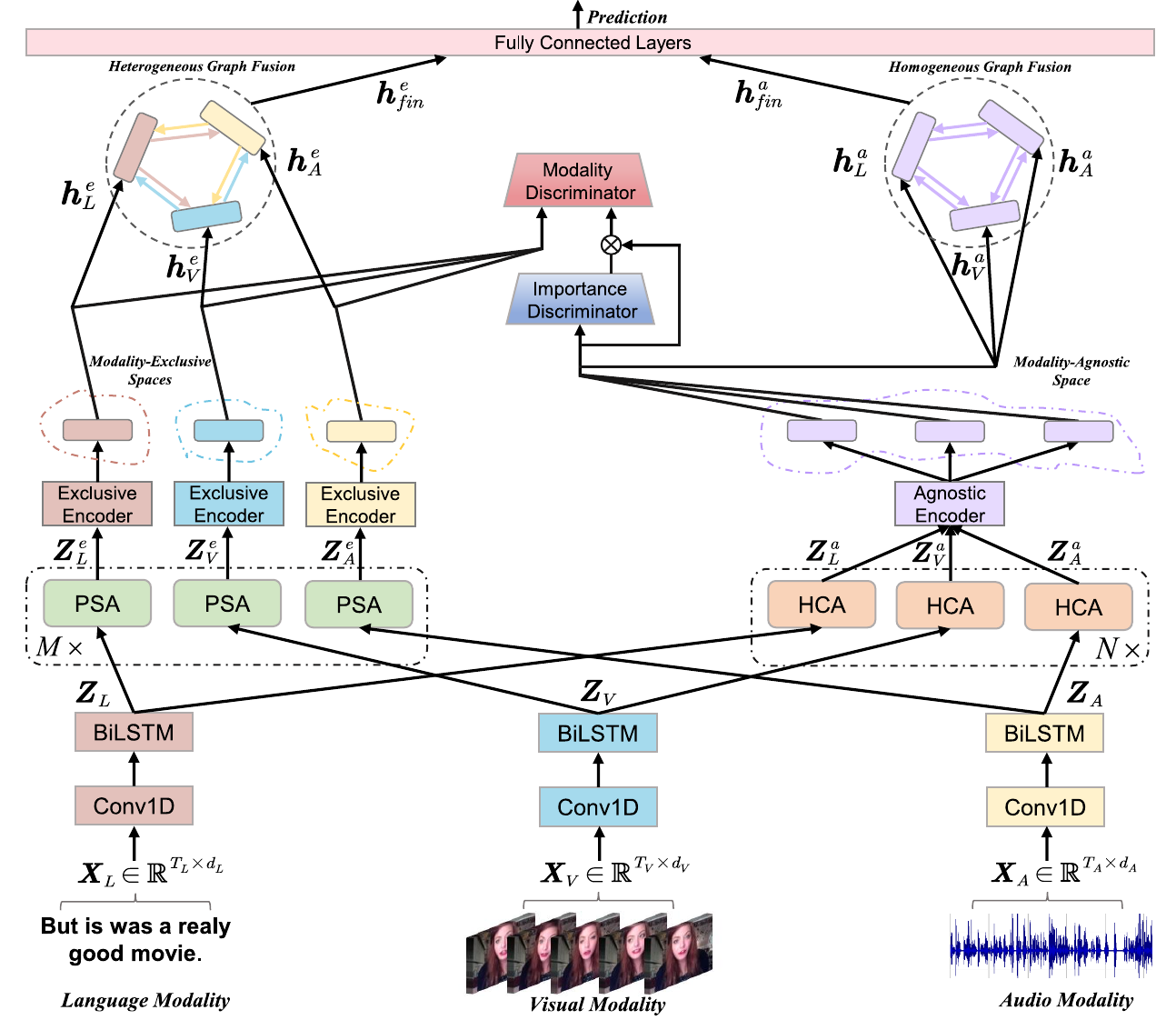}
  \caption{The overall architecture of the proposed Multimodal fusion approach for learning modality-Exclusive and modality-Agnostic representations (MEA). ``PSA" represents a predictive self-attention module. ``HCA" represents a hierarchical cross-modal attention module. 
}
  \label{arc}
\end{figure*}

We first outline the proposed MEA in this section. The overall model architecture is illustrated in Figure~\ref{arc}. 
Our goal is to perform asynchronous multimodal sequence fusion from three primary modalities extracted from video clips, including
Language ($L$), Visual ($V$), and Audio ($A$) modalities.  These pre-extracted multimodal sequences are denoted as 
$\bm{X}_L \in \mathbb{R}^{T_{L} \times d_{L}}$, $\bm{X}_V \in \mathbb{R}^{T_{V} \times d_{V}}$, and $\bm{X}_A \in \mathbb{R}^{T_{A} \times d_{A}}$, respectively, where $T_{(\cdot)}$ refers to the sequence length, and $d_{(\cdot)}$ represents the corresponding dimension of the feature embedding.

Firstly, we temporally preprocess the multimodal sequences to obtain the low-level representations $\bm{Z}_m \in \mathbb{R}^{T_{m} \times d}$, where $m \in \{ L, V, A\}$.
Afterward, two parallel branches are proposed to capture distinct representations from different modalities.
The first branch aims to employ the proposed Predictive Self-Attention (PSA) module to explicitly reinforce intra-modal dynamics and capture contextual dependencies with apriori knowledge.
Then, we proceed to project the enhanced features $\bm{Z}^{e}_m \in \mathbb{R}^{T_{m} \times d}$ from PSA into the modality-exclusive spaces employing three discrete exclusive encoders. This method aims to discern the distinctive attributes inherent to each modality and learn their intrinsic diversities.
The second branch focuses on cross-modal interactions at multiple granularities and capturing valuable element correlations of different modalities through the proposed Hierarchical Cross-modal Attention (HCA) module.
Then, a shared agnostic encoder projects the reinforced features $\bm{Z}^{a}_m \in \mathbb{R}^{T_{m} \times d}$ into the modality-agnostic space to learn the commonality and bridge the distribution gap among distinct modalities. 
In this case, a double-discriminator adversarial strategy is proposed to expressly supervise the generation of the modality-exclusive representations $\bm{h}^{e}_m \in \mathbb{R}^{ d_{h}}$ and the modality-agnostic representations $\bm{h}^{a}_m \in \mathbb{R}^{ d_{h}}$.
Further, we present a decoupled graph fusion mechanism to facilitate knowledge exchange and information sharing within $\bm{h}^{e}_m $  and $\bm{h}^{a}_m $.
Ultimately, the fused heterogeneous representation $\bm{h}^{e}_{fin} $  and homogeneous representation $\bm{h}^{a}_{fin} $ are fed into the fully connected layers for various downstream tasks.

\subsection{Uni-modal Extractor}
Uni-modal feature extraction provides a basic guarantee for subsequent feature refinement. Specifically, we first utilize a 1D temporal convolutional layer to preprocess the original multimodal sequences $\bm{X}_m \in \mathbb{R}^{T_{m} \times d_{m}}$, where $m \in \{ L, V, A\}$. This operation aims to compress the features of different modalities into the same dimension by controlling the size of the convolution kernel, denoted as $\bm{X}_m \in \mathbb{R}^{T_{m} \times d}$. We also augment the position embedding~\cite{vaswani2017attention} for each sequence to provide location-awareness capabilities. 
Subsequently, three separate Bi-directional Long Short Term Memory (Bi-LSTM) networks~\cite{hochreiter1997long} are deployed to extract preliminary characteristics of multimodal sequences:
\begin{equation}
    \bm{Z}_m = \text{Bi-LSTM}(\bm{X}_m ; \theta_{m}^{lstm} ) \in \mathbb{R}^{T_{m} \times d},
\end{equation}
where $\theta_{m}^{lstm}$ are the learnable parameters.

\begin{figure*}[t]
  \centering
  \includegraphics[width=0.95\textwidth]{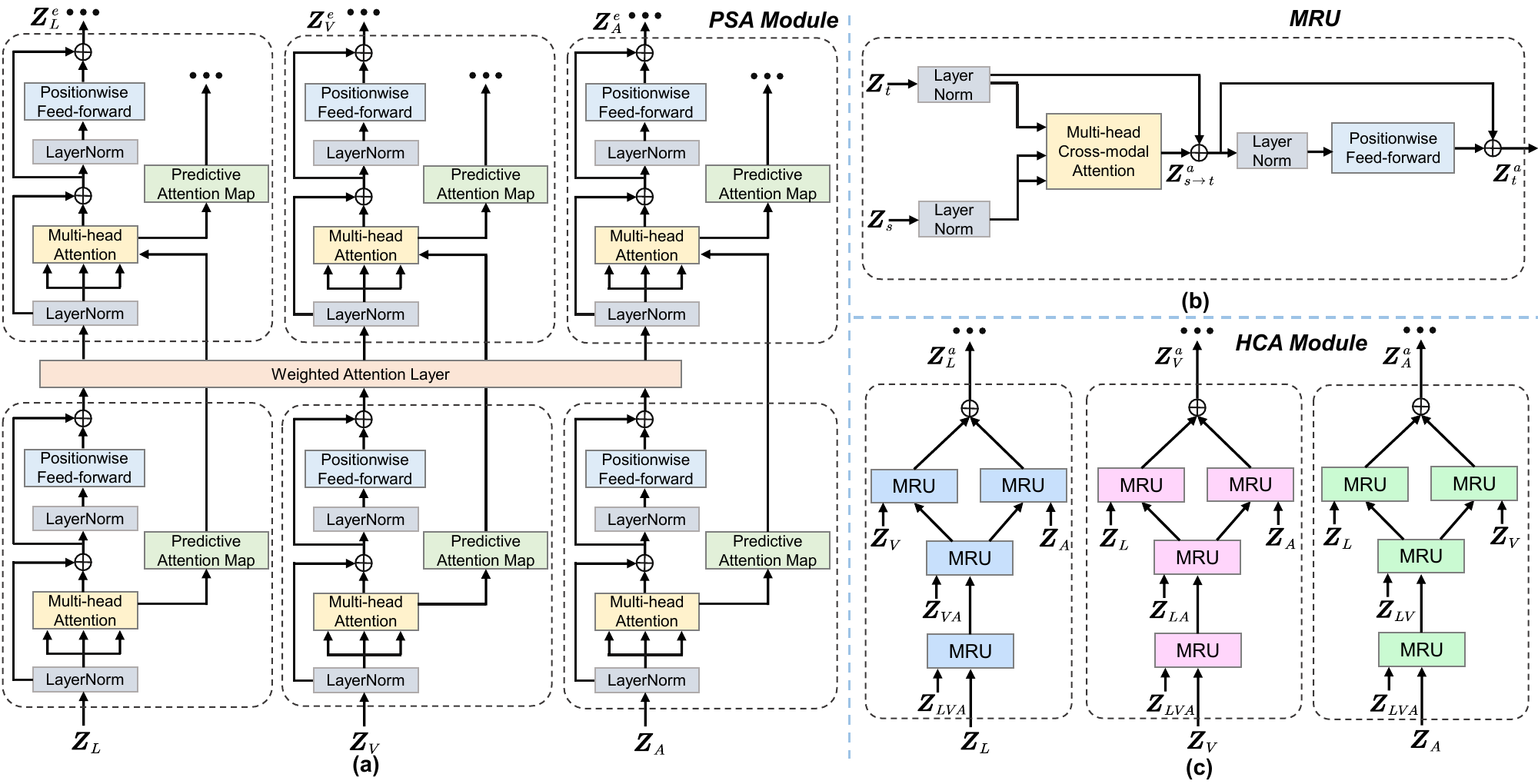}
  \caption{(a) The overall structure of the Predictive Self-Attention (PSA) modules. We provide a pipeline of two-layer PSA modules from three modalities to illustrate how the predictive attention map and weighted attention layer work.
  (b) The overall structure of a Modality Reinforcement Unit (MRU) in the HCA module. (c) The overall structure of the Hierarchical Cross-modal Attention (HCA) modules.}
  \label{arc_det}
\end{figure*}

\subsection{Predictive Self-Attention Module}
Transformers~\cite{vaswani2017attention} achieve exceptional performance in sequential modelling based on the attention-oriented interaction paradigm.
Nevertheless, previous works~\cite{yang2022learning,jain2019attention} have demonstrated the difficulty of a vanilla attention layer to capture reasonable long-range dependencies without the guidance of any apriori knowledge.
In addition, the original self-attention correlations of each level are learned independently, which limits the progressive enhancement of sequential representations from different modalities across low to high levels~\cite{wang2020predictive}.
Based on the above observations,
we design a Predictive Self-Attention (PSA) module to capture the intra-modal dynamics and reliable temporal dependencies.
Concretely, we design a convolution-driven prediction chain to compute the attention maps of the current module guided by the attention mapping of the preceding module.
The intuitive insight is that the chain would facilitate prior transfer and information interaction across different layers of attention patterns. As a result, the self-attention layer from each modality's current PSA module could incorporate modality-exclusive knowledge into the residual attention maps to provide valuable clues.

Figure~\ref{arc_det}(a) illustrates the two-layer stacked PSA modules from three modalities.
Following the vanilla attention~\cite{vaswani2017attention}, our PSA module contains \textit{Querys, Keys,} and \textit{Values}, denoted as
$\bm{Q}_m = LN(\bm{Z}_m) $ $ \bm{W}_{Q_m}$ with   $\bm{W}_{Q_m} \in \mathbb{R}^{d \times d}$, $\bm{K}_m = LN(\bm{Z}_m
)\bm{W}_{K_m} $ with   $\bm{W}_{K_m} \in \mathbb{R}^{d \times d}$, and $\bm{V}_m = LN(\bm{Z}_m) \bm{W}_{V_m} $ with   $\bm{W}_{V_m} \in \mathbb{R}^{d \times d}$, respectively, where $m \in \{ L,V,A\}$ and $LN$ means the layer normalization.
Given a mask matrix of attention logits denoted as $ A = \frac{\bm{Q}_m \bm{K}^{T}_m}{\sqrt{d}} \in \mathbb{R}^{T_{m} \times T_{m}}$. 
When performing a multi-attention operation with $K$ heads, we can naturally get $K$ attention logit maps.
Based on the shape of the total logit maps $\bm{A} \in \mathbb{R}^{T_{m} \times T_{m} \times K}$, we consider it as a $ T_{m} \times T_{m}$ intermediate feature with $K$ input channels. Thus, we adopt a 2D convolutional layer with $3 \times 3$ kernels to predict the attention maps for the subsequent module. The design philosophy is that the simple convolution operation can refine local attention dependencies without adding excessive overhead.
The predictive attention maps of all heads can be produced simultaneously by setting the number of output channels to remain $K$. Subsequently, a GeLU~\cite{hendrycks2016gaussian} function is employed to provide sparsity and non-linearity properties.
Ultimately, current dot product attention maps $\bm{A}_{cur}$ are improved by incorporating the previously predictive attention maps $\text{CNN}(\bm{A}_{pre})$, which is formulated as follows:
\begin{equation}
    \bm{A}=softmax(\mu \odot \text{CNN}(\bm{A}_{pre}) + (1-\mu ) \odot softmax(\bm{A}_{cur}) ),
\end{equation}
where $\odot$ represents that the multiplication operation follows the element-wise manner, and $\mu \in [0,1]$ is a hyper-parameter to balance the importance of two attention patterns. The first layer of the PSA module does not involve predictive attention maps. The subsequent computational procedure of the PSA module is formulated as follows:
\begin{align}
       \bm{Z}^{e}_{m} &= LN(\bm{Z}_{m})+ \bm{A} \bm{V}_{m}, \\
        \bm{Z}^{e}_{m} &= \mathcal{F}_{\theta}(LN(\bm{Z}^{e}_{m}))+\bm{Z}^{e}_{m},
\end{align}
where $ \mathcal{F}_{\theta} (\cdot)$ is a feed-forward network with the parameter $\theta$.
$\bm{Z}^{e}_{m}$is incrementally obtained and refined through the above attention and feed-forward transformation operations with residual connections.

In this branch, we aim to harvest the intra-modal dynamics and enhance the pure representation of each modality, serving the projection of the modality-exclusive spaces.
However, the heterogeneity observed across multiple modalities introduces surplus information within the domain of multimodal representations~\cite{yang2022disentangled}. 
To this end, we propose a Weighted Attention Layer (WAL) following the outputs from each layer of the PSA module across all three modalities to alleviate the impact of redundant information interference.
WAL dynamically improves the multimodal representations by adjusting the appropriate contribution of each modality using adaptive attention weights. Given the output $ \bm{Z}^{e}_{m} \in \mathbb{R}^{T_{m} \times d}$ of the PSA modules at any layer, we transform $ \bm{Z}^{e}_{m}$  into $ \tilde{\bm{Z}}^{e}_{m} \in \mathbb{R}^{T_{m}\cdot d \times 1}$ by a reshaping operation.
The attention weights can be adaptively obtained by the following formulation:
\begin{align}
       \gamma_{m} &= \bm{P}^{T}_{m} \cdot tanh(\bm{W}_{m}\cdot\tilde{\bm{Z}}^{e}_{m}+\bm{b}_{m}), \\
       \psi_{m} &= \frac{exp(\gamma_{m})}{\sum_{m\in \{ L,V,A\}}^{} exp(\gamma_{m})},
\end{align}
where $ \bm{P}_{m} \in \mathbb{R}^{T_{m}\cdot d \times 1}$ is a transformation vector, $\bm{W}_{m} \in \mathbb{R}^{T_{m}\cdot d \times T_{m}\cdot d} $ and $\bm{b}_{m} \in \mathbb{R}^{T_{m}\cdot d \times1} $ are the learnable parameters. The weighted multimodal representations are defined as $ \bm{Z}^{e}_{m} = \psi_{m} \odot \bm{Z}^{e}_{m}$. 
In practice, we minimally stack $M$-layer PSA modules to progressively enhance the multimodal representations $\bm{Z}^{e}_{m}$.

\subsection{Hierarchical Cross-Modal Attention Module}
Cross-modal interaction plays an essential role in asynchronous multimodal fusion.
Despite the significant improvements, existing methods either only considered shallow cross-modal adaptions~\cite{tsai2019multimodal,wang2023cross,sahay2020low,lv2021progressive} or focused on coarse-grained finite interactions~\cite{liang2021attention,yang2023target,lei2023text}, leading to learned element correlations among modalities that are potentially unreliable and indistinguishable.
To tackle this issue, 
we propose a Hierarchical Cross-modal Attention (HCA) module to capture meaningful inter-modal correlations through comprehensive cross-modal interactions.
The core philosophy is to achieve progressive adaptation transitions from source modalities $ \bm{Z}_s, s \in \{L, V, A \} $ to target modalities $ \bm{Z}_t, t \in \{L, V, A \} $ by seeking cross-modal representations.
We argue that cross-modal interactions can endow multimodal representations with superior modality adaptability that better serves the projection of the modality-agnostic space.

From Figure~\ref{arc_det}(c), the HCA module from each modality utilizes several Modality Reinforcement Units (MRUs) to explore cross-modal fusion at multiple granularities. During hierarchical feature interaction, multimodal representations accomplish semantic complementation and information exchange in a granularity-increasing paradigm.
To understand the working mechanism of our HCA, we start with the basic component MRU in Figure~\ref{arc_det}(b).
Formulaically speaking, we project the target modality as $\bm{Q}_t = LN(\bm{Z}_t) $ $ \bm{W}_{Q_t}$ with   $\bm{W}_{Q_t} \in \mathbb{R}^{d \times d}$ , and the source modality as $\bm{K}_s = LN(\bm{Z}_s) $ $ \bm{W}_{K_s}$ with   $\bm{W}_{K_s} \in \mathbb{R}^{d \times d}$ and $\bm{V}_s = LN(\bm{Z}_s) $ $ \bm{W}_{V_s}$ with   $\bm{W}_{V_s} \in \mathbb{R}^{d \times d}$.
The cross-modal adaptation is expressed as follows:
\begin{equation}
    \bm{Z}^{a}_{s \to t} = softmax(\frac{\bm{Q}_t \bm{K}^{T}_s}{\sqrt{d}})\bm{V}_s \in \mathbb{R}^{T_{t} \times d}.
\end{equation}
The subsequent forward calculation flow is represented as:
\begin{align}
       \bm{Z}^{a}_{t} &= LN(\bm{Z}_{t}) + \bm{Z}^{a}_{s \to t}, \\
        \bm{Z}^{a}_{t} &=  \mathcal{F}_{\delta }(LN( \bm{Z}^{a}_{t})) +\bm{Z}^{a}_{t},\, \bm{Z}^{a}_{t} \in \mathbb{R}^{T_{t} \times d},
\end{align}
where $ \mathcal{F}_{\delta} (\cdot)$ is the position-wise feed-forward layer with the parameter $\delta$. The procedure for a MRU is defined as $\bm{Z}^{a}_{t} = \text{MRU}(\bm{Z}_{s}, \bm{Z}_{t}) $.    
For clarity, we characterize the details of the HCA module using the language modality as a goal orientation.
The preliminary representations $ \bm{Z}_m, m \in \{ L,V,A\}$ from different modalities are concatenated at mixed and coarse granularities to produce  $ \bm{Z}_{LVA} = [\bm{Z}_{L}, \bm{Z}_{V}, \bm{Z}_{A} ] \in \mathbb{R}^{(T_{L}+T_{V}+T_{A}) \times d} $  and $ \bm{Z}_{VA} = [ \bm{Z}_{V}, \bm{Z}_{A} ] \in \mathbb{R}^{(T_{V}+T_{A}) \times d} $, respectively. $[\,\,,\,]$ stands for the concatenation operator.
In this case, the multi-grained cross-modal interaction and hierarchical feature fusion are summarized as follows:
\begin{align}
\text{Mixed-grained}: \hat{\bm{Z}}^{a}_{t} &= \text{MRU}(\bm{Z}_{LVA}, \bm{Z}_{L}), \nonumber \\
\text{Coarse-grained}: \check{\bm{Z}}^{a}_{t} &= \text{MRU}(\bm{Z}_{VA}, \hat{\bm{Z}}^{a}_{t}), \\
\text{Fine-grained}: \bm{Z}^{a}_{m} &= \text{MRU}(\bm{Z}_{V}, \check{\bm{Z}}^{a}_{t}) +  \text{MRU}(\bm{Z}_{A}, \check{\bm{Z}}^{a}_{t}). \nonumber 
\end{align}
During implementation, we minimally stack $N$-layer HCA modules to reinforce the multimodal representations $\bm{Z}^{a}_{m}$.

\subsection{Decoupled Representation Learning}
\subsubsection{Modality-Exclusive and -Agnostic Representations}
Learning informative multimodal representations is a critical component of asynchronous sequence fusion.
Previous SOTA efforts typically extracted indiscriminate representations from each modality, resulting in captured element correlations that could be ambiguous~\cite{tsai2019multimodal,lv2021progressive,lei2023text}. Another solution~\cite{liang2021attention} is removing noise interference across modalities in a latent space. Nevertheless, this one-sided strategy fails to consider the diversity of each modality, causing performance bottlenecks.
In contrast, our approach involves the acquisition of modality-exclusive and modality-agnostic representations for each modality, capitalizing on the synergistic information embedded within the amalgamation of multiple modalities.
The modality-exclusive representations attend to the distinctive characteristics of each modality and emphasize the diversity, which are designed upon the enhanced multimodal representations $\bm{Z}^{e}_{m}$ within the modality.
The modality-agnostic representations focus on eliminating distribution gaps among modalities and capturing the commonality, which are designed upon the refined multimodal representations $\bm{Z}^{a}_{m}$ across modalities.
Therefore, we devise three exclusive encoders for projecting  $\bm{Z}^{e}_{m}$ to the modality-exclusive spaces and an agnostic encoder for projecting $\bm{Z}^{a}_{m}$ to the modality-agnostic space:
\begin{align}
 \bm{h}^{e}_{m} &= \mathcal{S}_{m}(\bm{Z}^{e}_{m}; \theta_{m}) \in \mathbb{R}^{d_h}, \\
 \bm{h}^{a}_{m} &= \mathcal{A}(\bm{Z}^{a}_{m}; \theta_{\mathcal{A}}) \in \mathbb{R}^{d_h},
\end{align}
where $ \mathcal{S}_{m}(\cdot\,;\,\theta_{m})$ represent the exclusive encoders, which assign separate parameters  $\theta_{m}$ for each modality. $ \mathcal{A}(\cdot\,;\,\theta_{\mathcal{A}})$ represents the agnostic encoder, which shares the parameters $\theta_{\mathcal{A}}$ across all modalities. 
In practice, these encoders consist of two-layer perceptrons with the GeLU activation~\cite{hendrycks2016gaussian}.
\subsubsection{Disparity Constraint}
To encourage the exclusive and agnostic encoders to distinguish distinct representations and penalize semantic redundancy across different modalities, we introduce the Hilbert-Schmidt Independence Criterion (HSIC)~\cite{song2007supervised} to efficiently measure the independence between decoupled representations. 
The intuition is that if the independence between two representations is high, their difference is significant.
Formulaically speaking, the HSIC disparity constraint between any two decoupled representations is expressed as:
\begin{equation}
   \text{HSIC} ( \boldsymbol{h}_{m}^{e}, \boldsymbol{h}_{m}^{a}) = (n-1)^{-2} Tr(\boldsymbol{U}\boldsymbol{K}_{m}^{e}\boldsymbol{U}\boldsymbol{K}_{m}^{a}).
\end{equation}
Here, $\boldsymbol{K}_{m}^{e}$ and $\boldsymbol{K}_{m}^{a}$ are the Gram matrices with $k_{m, ij}^{e} = k_{m}^{e} $ $(\boldsymbol{h}_{m}^{i,e},\boldsymbol{h}_{m}^{j,e})$ and $k_{m, ij}^{a} = k_{m}^{a}(\boldsymbol{h}_{m}^{i,a},\boldsymbol{h}_{m}^{j,a} )$. $\boldsymbol{U}=\boldsymbol{I}- (1/n) ee^{T} $, where $\boldsymbol{I}$ is an identity matrix and $e$ is an all-one column vector.
In our implementation, the inner product kernel is used for $\boldsymbol{K}_{m}^{e}$ and $\boldsymbol{K}_{m}^{a}$.
The HSIC constraint is computed among the representations associated with each modality pair, and the collective disparity constraint is formally articulated as follows:
\begin{equation}
  \mathcal{L}_{dis} = \frac{1}{3}\sum_{m \in \{ L, A, V  \}}^{} \text{HSIC} ( \boldsymbol{h}_{m}^{e}, \boldsymbol{h}_{m}^{a}).
\end{equation}
\subsubsection{Double-Discriminator Adversarial Strategy}

To ensure that $\bm{h}^{e}_{m}$ exactly emphasizes the specific properties of each modality and $\bm{h}^{a}_{m}$ depicts the informational commonalities among different modalities, we propose a double-discriminator adversarial strategy to supervise the parameter optimization of the exclusive and agnostic encoders.
We formulate the ground truths of the modality labels of $\bm{h}^{e}_{m}$ and $\bm{h}^{a}_{m}$, which are expressed as $y_L = [1,0,0],\, y_V= [0,1,0],$ and $y_A = [0,0,1]$.
From the upper part of Figure~\ref{arc}, the first subcomponent is an importance discriminator, considered as a classifier formulated as $\mathcal{D}_{i}(\bm{h}^{a}_{m};\theta_{\mathcal{D}_i}) = softmax((\bm{h}^{a}_{m})^{T}\cdot \bm{W}_{i})$.
$\bm{W}_{i} \in \mathbb{R}^{d_{h} \times 3}$ is the learnable matrix. If $\bm{h}^{a}_{m}$ is accurately identified as belonging to modality $m$, $\mathcal{D}_{i}(\cdot\,;\,\theta_{\mathcal{D}_i})$ would converge to the possibly optimal solution.
Thus, the importance discriminator gives the likelihood of $\bm{h}^{a}_{m}$ based on the discriminative modality $m$. 
When $\mathcal{D}_{i}(\bm{h}^{a}_{m};\theta_{\mathcal{D}_i}) \approx 1 $, $\bm{h}^{a}_{m}$ contains very few modality-agnostic semantics among different modalities since it can be entirely distinguished from other modalities.
In this case, we treat the degree of $\bm{h}^{a}_{m}$ as a regulatory factor $\omega^{a}_{m}$ to imply contributing to the modality-agnostic representations. $\omega^{a}_{m}$ is defined as $\omega^{a}_{m} =1- \mathcal{D}_{i}(\bm{h}^{a}_{m};\theta_{\mathcal{D}_i})$ because it should be inversely related to $\mathcal{D}_{i}(\bm{h}^{a}_{m};\theta_{\mathcal{D}_i})$.

The second subcomponent is a modality discriminator $\mathcal{D}_{m}(\mathcal{H};\theta_{\mathcal{D}_m})$ that transforms the input $\mathcal{H}$ into an estimated probability distribution and facilitates the disentanglement of distinct representations. Here, the input $\mathcal{H}$ comes either from the output $\bm{h}^{e}_{m}$ of the exclusive encoders or from the output $\bm{h}^{a}_{m}$ of the agnostic encoder.

By incorporating the degree $\omega^{a}_{m}$ when the discriminator $\mathcal{D}_{m}(\cdot\,;\,\theta_{\mathcal{D}_m})$ is supervised over the modality-agnostic representations, the agnostic adversarial constraint is expressed as:
\begin{equation}
    \mathcal{L}_{agn} = - \frac{1}{n}\sum_{m}^{}  \sum_{k=1 }^{n} (y_m \omega^{a}_{m}log(\mathcal{D}_{m}(\bm{h}^{a}_{m};\theta_{\mathcal{D}_m}))),
\end{equation}
where $m \in \{ L,V,A \}$ and $n$ is the sample count in a batch. In practice, we augment a gradient reversal layer~\cite{ganin2015unsupervised} to achieve the local optimization of $\mathcal{L}_{agn}$.

To capture modality-exclusive representations in different projection spaces, the modality discriminator is also employed to distinguish the source of different modalities. 
The exclusive adversarial constraint is formulated as follows:
\begin{equation}
    \mathcal{L}_{exc} = - \frac{1}{n}\sum_{m}^{}  \sum_{k=1 }^{n} (y_m log(\mathcal{D}_{m}(\bm{h}^{e}_{m};\theta_{\mathcal{D}_m}))).
\end{equation}

\subsection{Decoupled Graph Fusion Mechanism}
The modality heterogeneity dilemma leads to pronounced differences in knowledge intensity and semantic information from different modalities~\cite{yu2021learning}. In order to bridge the inter-modality gap, we present a Decoupled Graph Fusion (DGF) mechanism to enhance the decoupled representations of each modality. DGF combines heterogeneous and homogeneous graph fusion to integrate modality-exclusive and -agnostic representations. We describe the details in terms of heterogeneous graph fusion as an example.
In the directed heterogeneous graph $\mathcal{G}$, $\bm{h}^{e}_{i}$ denotes the $i$-th node, and $\delta_{i,j}$ denotes the semantic strength from the $j$-th neighbouring node, which is computed as follows:
\begin{equation}
    \delta_{i,j} = \mathcal{Q} ([\bm{W}_{e}\bm{h}^{e}_{i}, \bm{W}_{e}\bm{h}^{e}_{j}]; \theta_{\mathcal{Q}}),\,i,j \in \{ L, V, A \},\,j \in \mathcal{N}_{i},
\end{equation}
where $\mathcal{Q} (\cdot\,;\,\theta_{\mathcal{Q}})$ indicates a single-layer feed-forward network parametrized by $\theta_{\mathcal{Q}}$, and $\bm{W}_{e} \in \mathbb{R}^{d_h \times d_h}$ is a shared projection matrix.
Immediately, the inter-modality  knowledge transfer coefficient is defined as follows:
\begin{equation}
    \xi_{i,j} = \frac{exp(\text{GeLU}(\delta_{i,j}))}{ {\textstyle \sum_{k\in\mathcal{N}_{i}}^{}} exp(\text{GeLU}(\delta_{i,k}))}.
\end{equation}
The exclusive representations from each modality are aggregated to produce the enhanced representation:
\begin{equation}
 \bm{h}^{e}_{fin} = \sum_{i\in\{L, V, A \}}^{} \sigma(\sum_{j\in\mathcal{N}_{i}}^{}\xi_{i,j}\odot\bm{W}_{e}\bm{h}^{e}_{j}),
\end{equation}
where $\sigma$ denotes the sigmoid activation. Similarly, the reinforced agnostic representation $\bm{h}^{a}_{fin}$ is obtained by homogeneous graph fusion following the procedure described above. We eventually concatenate $\bm{h}^{e}_{fin}$  and $\bm{h}^{a}_{fin}$  and make predictions via the fully connected layers.

\subsection{Objective Optimization}

The task-related losses are selected to follow the consensus of previous mainstream works~\cite{yang2022disentangled,tsai2019multimodal,Hazarika2020mm} for a fair comparison.
We use the standard cross-entropy loss $ \mathcal{L}_{task} = - \frac{1}{n} \sum_{k=1}^{n} y_k \cdot log \hat{y}_k $ for the classification task.
For the regression task, we utilize the mean squared error loss $ \mathcal{L}_{task} = \frac{1}{n} \sum_{k=1}^{n} \parallel y_k - \hat{y}_k \parallel^{2}_{2} $, where $y_k$ is the ground truth and $\hat{y}_k$ is the predicted out. Combining the task loss $\mathcal{L}_{task}$, disparity loss $\mathcal{L}_{dis}$, and adversarial losses $\mathcal{L}_{agn}\,\&\,\mathcal{L}_{exc}$, the total loss is expressed as:
\begin{equation}
    \mathcal{L}_{all} = \mathcal{L}_{task}+ \alpha \mathcal{L}_{dis} +  \beta (\mathcal{L}_{agn}+ \mathcal{L}_{exc}),
\end{equation}
where $\alpha$ and $\beta$ are the trade-off coefficients.

\begin{table}[t]
\setlength{\tabcolsep}{9pt}
\centering
\caption{We show the hyper-parameter settings on the MOSI, MOSEI, and IEMOCAP datasets. All the hyper-parameters are determined via the validation set of each dataset.}
\resizebox{\linewidth}{!}{%
\begin{tabular}{c|ccc}
\toprule
Setting               & MOSI  & MOSEI & IEMOCAP \\ \midrule
Batch  Size          & 32    & 64    & 32      \\
Epoch  Number        & 60    & 100   & 60      \\
Hidden  Dimension $d$    & 40    & 40    & 40      \\
Output  Dimension  $d_h$   & 64    & 64    & 64      \\
Learning  Rate       & 1e-3  & 2e-3  & 1e-3    \\
Attention  Head      & 8     & 10    & 8       \\
Kernel Size (L/V/A)         & 3/3/3 & 3/3/3 & 3/3/5   \\
Coefficient $\mu$       & 0.25  & 0.15  & 0.2     \\
Trade-off  Parameter $\alpha$ & 2e-2  & 3e-2  & 1e-2    \\
Trade-off  Parameter $\beta$ & 3e-2  & 5e-2  & 2e-2    \\ \bottomrule
\end{tabular}
}
\label{tab_setting}
\end{table}

\section{Datasets and Implementation Details}
\label{sec4}

Extensive experiments are conducted on three multimodal sequence fusion datasets,
including MOSI~\cite{zadeh2016multimodal}, MOSEI~\cite{zadeh2018multimodal}, and IEMOCAP~\cite{busso2008iemocap}.
All used benchmarks are publicly available.
These datasets focus on human sentiment analysis and emotion recognition, which provide asynchronous sequences from different modalities for each sample.
We follow the public protocol consistent with previous SOTA works~\cite{tsai2019multimodal, lv2021progressive, liang2021attention} for fair comparisons.

\subsection{Datasets and Evaluation Metrics}

\textbf{MOSI}~\cite{zadeh2016multimodal} is a multimodal sentiment analysis dataset consisting of 2,199 video clips. These clips contain short monologues from different subjects commenting on social media.
The standard partitioning of the dataset is 1,284 samples in the training set, 229 in the validation set, and 686 in the testing set.
The linguistic data is segmented by words and represented as discrete word embeddings. 
The acoustic features are obtained at a sampling frequency of 12.5 Hz, while the visual features have a processing resolution of 15 Hz. Each multimodal sample is manually annotated with a continuous sentiment score that ranges from -3 to 3. The higher score represents the more positive sentiment intensity of the subject.

\noindent \textbf{MOSEI} \cite{zadeh2018multimodal} is a large-scale sentiment analysis benchmark containing 22,856 samples. These samples are captured from movie review clips with different content sources.
The predefined dataset division is summarized as 16,326 training samples, 1,871 validation samples, and 4,659 testing samples. The language modality is segmented per word and then transformed into the corresponding word embedding. The acoustic and visual features are characterized based on sampling ratios of 20 Hz and 15 Hz, respectively. Each data sample is assigned an opinion score between [-3, 3] to represent a variation in sentiment polarity from strongly negative to strongly positive.

To thoroughly evaluate our approach, we adopt various metrics on the MOSI and MOSEI datasets, including $Acc_7$: 7-class accuracy of sentimental polarity classification, $Acc_2$: binary accuracy of positive/negative emotions, $F1$ score, $MAE$: mean absolute error, and $Corr$: the correlation of the model’s prediction with human.

\begin{table}[t]
\setlength{\tabcolsep}{5pt}
\centering
\caption{Comparison of testing set results on the MOSI dataset. Best results are marked in \textbf{bold}. $\dagger$ means the results are reproduced from the available codebase. $\sharp$ means the results from our conference version. $\flat $ means the corresponding results are significantly better than SOTA with p-value !` 0.05 based on paired t-test. These footnotes to Tables~\ref{tab_mosei} and~\ref{tab_iemocap} follow identical interpretations.}
\resizebox{\linewidth}{!}{%
\begin{tabular}{cccccc}
\toprule
Approach   & $Acc_7 \uparrow$ & $Acc_2 \uparrow$ & $F1 \uparrow$ & $MAE \downarrow$   & $Corr \uparrow$  \\ \midrule
EF-LSTM    &31.0 &73.6 &74.5 &1.078 &0.542 \\ 
LF-LSTM    &33.7 &77.6 &77.8 &0.988 &0.624 \\ 
RAVEN  \cite{wang2019words}  &31.7 &72.7 &73.1 &1.076 &0.544 \\ 
MCTN \cite{pham2019found} & 32.7 &75.9 & 76.4 &0.991 & 0.613 \\ 
MulT \cite{tsai2019multimodal}  & 39.1 & 81.1 & 81.0 & 0.889 & 0.686 \\ 
HCT-MG \cite{wang2023cross}  & 39.4	&  82.5 & 82.5 & 0.881 & 0.710 \\ 
PMR \cite{lv2021progressive}  & 40.6 & 82.4 & 82.1& - & - \\ 
MICA \cite{liang2021attention}  & 40.8 & 82.6 & 82.7 & - & - \\ 
Self-MM$^{\dagger}$ \cite{yu2021learning}  & 41.2 & 82.9 & 83.2	& 0.863	& 0.717\\ 
MMIM$^{\dagger}$ \cite{han2021improving}  & 41.7 & 83.6 & 83.5 & 0.854 & 0.725 \\ 
DMD \cite{li2023decoupled}  & 41.9 & 83.5 & 83.5 & - & - \\ 
MFSA$^{\sharp}$ (ours) \cite{yang2022learning}  & 41.4 & 83.3 & 83.7 & 0.856 & 0.722 \\ 
\textbf{MEA (ours)} & \textbf{42.5}$^{\flat}$ & \textbf{84.4}$^{\flat}$ & \textbf{84.6}$^{\flat}$ & \textbf{0.844}$^{\flat }$ & \textbf{0.735}$^{\flat}$ \\ \bottomrule
\end{tabular}%
}
\label{tab_mosi}
\vspace{-3pt}
\end{table}

\noindent \textbf{IEMOCAP}~\cite{busso2008iemocap} is a conversational emotion recognition dataset containing language, audio, and visual modalities. The data samples are recorded by a motion capture camera from scripted videos of 10 actors. 
The audio signals are sampled at a fixed rate of 12.5 Hz, while the visual signals are recorded at a resolution of 15 Hz.
We follow \cite{wang2019words}'s recommendation to perform experiments using four emotion labels (\ie, happy, sad, angry, and neutral) to ensure a fair comparison with most works.
The classification accuracy ($Acc$) and $F1$ score are utilized as evaluation metrics for each emotion category.

\subsection{Implementation Details}
To fairly and intuitively evaluate the proposed MEA, all methods utilize the \textit{same feature extraction procedures} on asynchronous raw data.
For the language modality, we transform the transcript texts from each video into pre-trained Glove-based word embedding~\cite{pennington2014glove} and characterize it as the 300-dimensional vector.
For the visual modality, the mainstream Facet~\cite{baltruvsaitis2016openface} instrument is employed to inscribe 35 facial action units, which record emotion-related facial movement clues.  In addition, we utilize the COVAREP toolkit~\cite{degottex2014covarep} to capture low-level acoustic characteristics having 74 semantic dimensions.
These characteristics include voiced segmenting features, 12 Mel-Frequency Cepstral Coefficients (MFCCs), and glottal source parameters. All models are implemented over the Pytorch toolbox, and the computational resources are two Quadro RTX 8000 GPUs. The Adam optimizer~\cite{kingma2014adam} is employed for model optimization. 
Based on the preliminary feature embeddings for different modalities, we jointly train all the components in the proposed framework in an end-to-end manner.
We provide the hyper-parameter configurations on each dataset in Table~\ref{tab_setting}, including the batch size, epoch number, feature dimension, learning rate, kernel size, and so on. The kernel size in the 1D temporal convolutional layer is utilized to align the feature dimensions from each modality during the preprocessing stage. Moreover, we minimally use 3-layer (\ie, $M=3$) PSA modules and 2-layer (\ie, $N=2$) HCA modules separately to reduce the model complexity. All the hyper-parameters are determined via the corresponding validation set in each dataset.

\begin{table}[t]
\setlength{\tabcolsep}{5pt}
\centering
\caption{Comparison of testing set results on the MOSEI dataset. Best results are marked in \textbf{bold}.}
\resizebox{\linewidth}{!}{%
\begin{tabular}{cccccc}
\toprule
Approach         & $Acc_7 \uparrow$ & $Acc_2 \uparrow$ & $F1 \uparrow$ & $MAE \downarrow$   & $Corr \uparrow$           \\ \midrule
EF-LSTM    &46.3 &76.1 &75.9 &0.680 &0.585 \\ 
LF-LSTM    &48.8 &77.5 &78.2 &0.624 &0.656 \\ 
RAVEN  \cite{wang2019words}  &45.5 &75.4 &75.7 &0.664 &0.599 \\
MCTN \cite{pham2019found} &48.2 &79.3 &79.7 &0.631 &0.645 \\ 
MulT   \cite{tsai2019multimodal}  & 50.7  & 81.6   &81.6 & 0.591 & 0.694 \\ 
HCT-MG \cite{wang2023cross}  & 50.6 &81.8&81.9&	0.593&	0.691 \\ 
PMR \cite{lv2021progressive}  & 51.8 & 83.1 & 82.8& - & - \\ 
MICA \cite{liang2021attention}  & 52.4 & 83.7 & 83.3 & - & - \\ 
Self-MM$^{\dagger}$ \cite{yu2021learning}  & 52.7&83.8&	83.6&0.579 &	0.723\\ 
MMIM$^{\dagger}$ \cite{han2021improving}  & 53.5&84.2&	84.2	&0.571&	0.726 \\ 
DMD \cite{li2023decoupled}  & 54.6	& 84.8 &	84.7 & - & - \\ 
MFSA$^{\sharp}$ (ours) \cite{yang2022learning}  & 53.2	&83.8	&83.6	& 0.574 & 0.724 \\ 
\textbf{MEA (ours)} & \textbf{54.8}$^{\flat}$ & \textbf{85.2}$^{\flat}$ & \textbf{85.1}$^{\flat}$ & \textbf{0.563}$^{\flat}$  & \textbf{0.731}$^{\flat}$ \\ \bottomrule
\end{tabular}%
}
\label{tab_mosei}
\end{table}

\begin{table*}[t]
\setlength{\tabcolsep}{15pt}
\centering
\caption{Comparison of testing set results on the IEMOCAP dataset. Best results are marked in \textbf{bold}. }
\label{tab_iemocap}
\resizebox{\linewidth}{!}{%
\begin{tabular}{ccccccccc}
\toprule
Category & \multicolumn{2}{c}{Happy} & \multicolumn{2}{c}{Sad} & \multicolumn{2}{c}{Angry} & \multicolumn{2}{c}{Neutral} \\ \cmidrule{1-9} 
Approach          & $ Acc \uparrow$  & $ F1 \uparrow$   & $ Acc\uparrow$  & $ F1\uparrow$    & $ Acc\uparrow$  & $ F1\uparrow$    & $ Acc\uparrow$  & $ F1\uparrow$    \\ \midrule
EF-LSTM    &76.2 &75.7 &70.2 &70.5 &72.7 &67.1 &58.1 &57.4 \\
LF-LSTM    &72.5 &71.8 &72.9 &70.4 &68.6 &67.9 &59.6 &56.2 \\
RAVEN \cite{wang2019words}    &77.0 &76.8 &67.6 &65.6 &65.0 &64.1 &62.0 &59.5 \\
MCTN  \cite{pham2019found}      &80.5 &77.5 &72.0 &71.7 &64.9 &65.6 &49.4 &49.3 \\
MulT  \cite{tsai2019multimodal}    & 84.8 & 81.9 & 77.7 & 74.1 & 73.9 & 70.2 & 62.5 & 59.7 \\
HCT-MG \cite{wang2023cross}    & 85.6 & 79.0 & 79.4 & 70.3 & 75.8 & 65.4 & 61.0 & 50.5 \\
PMR \cite{lv2021progressive} & 86.4 & 83.3 &78.5 & 75.3& 75.0 & 71.3& 63.7 &60.9 \\
MICA \cite{liang2021attention}  & 86.8 & 83.9 & 79.3 & 75.2 & 75.7 & 72.4 & 63.7 & 61.6 \\
Self-MM$^{\dagger}$ \cite{yu2021learning}  & 86.5 & 83.5 & 78.7 & 75.4 & 75.2 & 71.9 & 63.5 & 60.7 \\
MMIM$^{\dagger}$ \cite{han2021improving}  & 87.0 & 84.1 & 80.5 & 76.4 & 76.2 & 72.8 & 64.6& 62.8 \\
DMD$^{\dagger}$ \cite{li2023decoupled}  & 87.0 & 84.1 & 79.6 & 75.5 & 75.1 & 71.6 & 63.8 & 62.2 \\
MFSA$^{\sharp}$ (ours)  & 87.2 & 84.3 & 80.7 & 76.8 & 76.5 & 73.2 & 64.4 & 62.5 \\
\textbf{MEA (ours)} & \textbf{87.5}$^{\flat}$ & \textbf{84.9}$^{\flat}$ & \textbf{81.8}$^{\flat}$ & \textbf{77.2}$^{\flat}$ & \textbf{76.8}$^{\flat}$ & \textbf{73.7}$^{\flat}$ & \textbf{65.6}$^{\flat}$ & \textbf{63.0}$^{\flat}$ \\ \bottomrule
\end{tabular}%
}
\end{table*}

\section{Experimental Results and Analysis}
\label{sec5}
\subsection{Comparison with State-of-the-Art Methods}
Despite numerous current works that have been evaluated on selected datasets, they have significant differences in language feature preprocessing and word alignment operations. Considering the unaligned setup of this paper, we mainly focus on comparing the proposed MEA with the SOTA efforts that directly deal with asynchronous multimodal sequences to ensure intuitiveness and fairness. These comparable approaches include Late Fusion LSTM (LF-LSTM), Multimodal Transformer (MulT) \cite{tsai2019multimodal}, Hierarchical Crossmodal Transformer with Modality Gating (HCT-MG)~\cite{wang2023cross}, Progressive Modality Reinforcement (PMR) \cite{lv2021progressive}, Modality-Invariant Crossmodal Attention (MICA)\cite{liang2021attention},
Self-Supervised Multi-task Multimodal Network (Self-MM)~\cite{yu2021learning}, MultiModal InfoMax (MMIM)~\cite{han2021improving}, and Decoupled Multimodal Distillation (DMD)~\cite{li2023decoupled}.
Among these models, we reproduce the results of Self-MM and MMIM following the Glove-based word embedding to maintain fair comparisons with most methods.

For a more holistic comparison, we augment a Connectionist Temporal Classification (CTC) loss \cite{graves2006connectionist} for several prominent models that cannot directly handle asynchronous sequence fusion, including Early Fusion LSTM (EF-LSTM), Recurrent Attended Variation Embedding Network (RAVEN)~\cite{wang2019words}, and Multimodal Cyclic Translation Network (MCTN)~\cite{pham2019found}. Specifically, these models simultaneously optimize the task-relevant objective and CTC alignment supervision during training.

The experimental results on the MOSI, MOSEI, and IEMOCAP datasets are reported in  Tables~\ref{tab_mosi}, \ref{tab_mosei}, and \ref{tab_iemocap}, respectively. We have the following key observations.

(i) The proposed model significantly outperforms the previous methods~\cite{tsai2019multimodal,lv2021progressive,liang2021attention,wang2023cross,yu2021learning,li2023decoupled,han2021improving} without additional data alignment on the three datasets, revealing the merit of MEA in asynchronous sequence fusion.
We note that existing methods either report only incomplete results (\eg, PMR, MICA, and DMD are missing evaluations on the $MAE$ and $Corr$ metrics for the MOSI and MOSEI datasets) or improve on only partial metrics (\eg, DMD is lower than our previous version of the MFSA on the $F1$ score for the MOSI dataset). These inadequate assessments highlight performance bottlenecks in the previous models. In contrast, MEA shows significant performance gains on all metrics. The reasonable interpretation is that MEA mitigates the information redundancy and distribution gaps across modalities in an agnostic space while refining the exclusive characteristics of each modality, yielding a more comprehensive improvement. Compared with our previous MFSA, the newly designed decoupled graph fusion mechanism further alleviates the modality heterogeneity dilemma and enables our MEA to achieve a new SOTA.

(ii) Our MEA evidently provides about 8$\sim$12\% improvements in most attributes over works~\cite{wang2019words,pham2019found} that require CTC loss to perform the temporal alignment.
This finding indicates the inability of traditional word-alignment-based methods to cope with the challenges of asynchronous sequence fusion. In comparison, our approach effectively learns intra-modal dynamics and cross-modal interactions via the tailored PSA and HCA modules, respectively, which may have superior advantages in capturing element context dependencies on long-range asynchronous sequences.

(iii) From the representation learning perspective, MEA not only outperforms efforts~\cite{tsai2019multimodal,wang2019words,pham2019found,lv2021progressive} that do not distinguish the hybrid representations of each modality, but also is superior to the research~\cite{liang2021attention} that learns element correlations only in the modality-invariant space. The above observations demonstrate that it is beneficial to consider both modality-exclusive and modality-agnostic representations in multimodal sequence fusion. The diversity of each modality in the exclusive spaces and the commonality among modalities in the agnostic space potentially form complementary semantics to yield more robust multimodal representations.

\subsection{Visualization and Qualitative Evaluation}
Here, we provide systematic visualizations to qualitatively verify the necessity of the proposed components in MEA.

\subsubsection{Visualization of PSA Module}
\begin{figure}[t]
  \centering
  \includegraphics[width=0.9\linewidth]{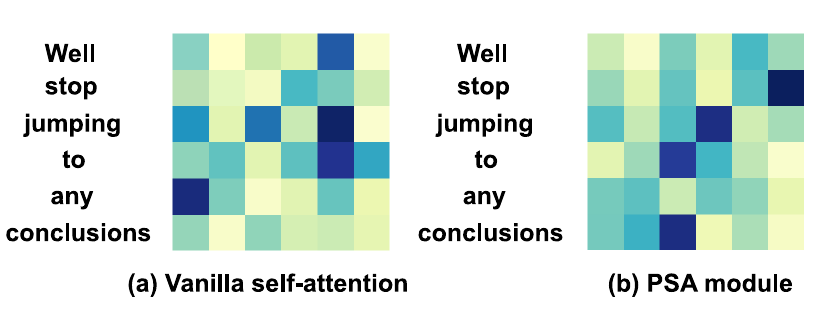}
  \caption{We show the attention matrix activations from (a) the vanilla self-attention \cite{vaswani2017attention} and (b) the proposed PSA module in the language modality. Compared to vanilla self-attention, our PSA module captures more meaningful attention correlations across elements within the modality.}
  \label{vis_PSA}
\end{figure}

To prove the merit of our model in learning contextual dependencies and intra-modal dynamics, we randomly visualize the attention activations of the language modality from a sample on the MOSEI dataset. The attention activations come from the last layer of the vanilla self-attention module~\cite{vaswani2017attention} and the proposed PSA module separately to provide an intuitive comparison.
From Figure~\ref{vis_PSA}(a), the vanilla self-attention focuses on confusing connections between spoken words, such as ``well'' and ``any''.
These attention associations are meaningless due to the inability to reflect the negative emotion aligned with the sample's ground truth.
In contrast, our PSA module attends to the valuable relationship between ``stop" and ``conclusions", which correctly captures the specific context semantics (\ie, predicate verb phrase) of the language modality in Figure~\ref{vis_PSA}(b).
These observations indicate that the attention pattern incorporating convolutional induction bias facilitates semantic refinement within modalities. The finding may benefit from the projection constraint when learning the modality-exclusive representations.

\begin{figure*}[t]
  \centering
  \includegraphics[width=\textwidth]{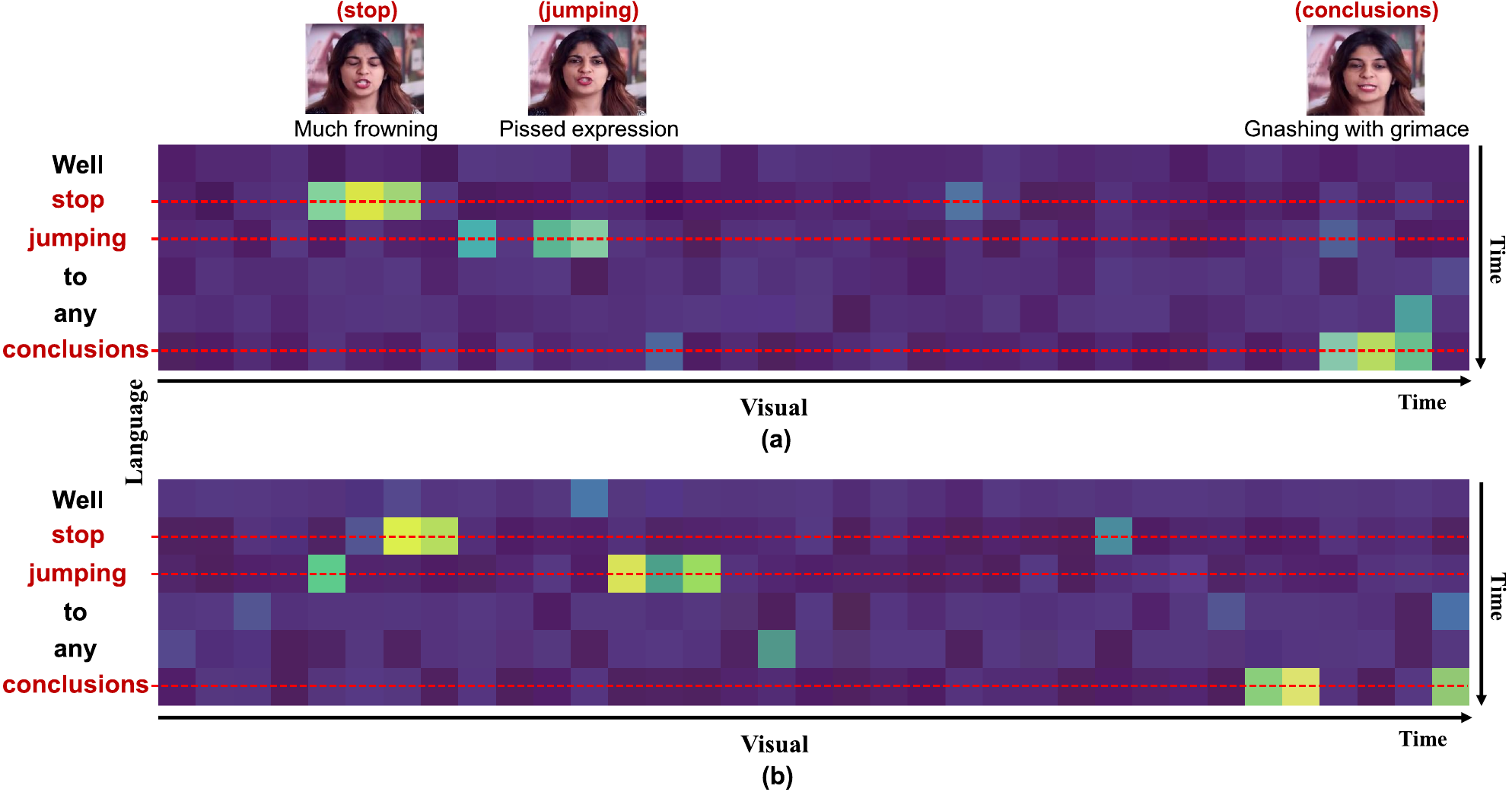}
  \caption{We show the cross-modal attention matrix activations of (a) the proposed HCA module and (b) the SOTA method DMD~\cite{li2023decoupled} on the MOSEI dataset. The spoken words closely related to the expression of human emotions are marked in red. Compared to the DMD, our model learns more reliable element correlations between different modalities. For example, stronger attention weights are focused on the intersection regions of cross-modal elements on asynchronous sequences between spoken words (``conclusions") and video frames (``gnashing with grimace''), which usually suggest salient emotion clues.}
  \label{vis_HCA}
\end{figure*}

\begin{figure*}[t]
  \centering
  \includegraphics[width=\textwidth]{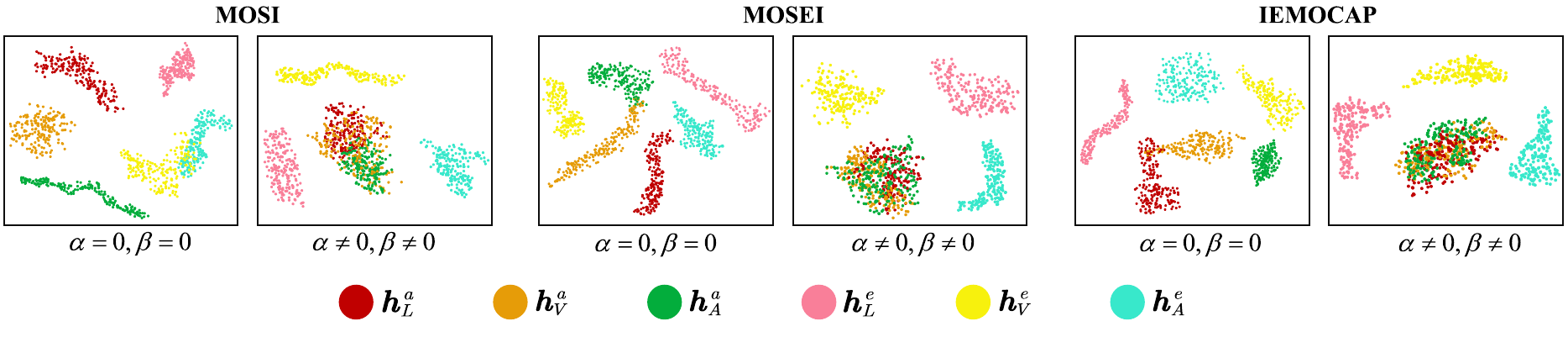}
  \caption{We randomly select 200 samples in the testing set on the three datasets to visualize modality-agnostic and -exclusive representations. $\alpha = 0, \beta =0 $ denotes without disparity and adversarial constraints, and vice versa. The red, orange, and green colours correspond to agnostic parts. The pink, yellow, and blue colours correspond to the exclusive parts.}
  \label{vis_decoplued}
\end{figure*}

\subsubsection{Visualization of HCA Module}
For consistency, the same sample on the MOSEI dataset is selected to evaluate the proposed HCA module's ability to model cross-modal interactions. Figure~\ref{vis_HCA}(a)\&(b) show the attention matrix activations for the last layer of the fine-grained MRU in our HCA module and the multimodal transformer in the SOTA DMD (CVPR2023)~\cite{li2023decoupled}, respectively. The brighter regions imply stronger attention dependencies between cross-modal elements. We observe that our module captures reasonable affective clues between visual elements in video frames and linguistic elements in spoken words. For instance, the emotion-related words (\eg, ``stop'' and ``conclusions'') successfully attend to the video frames that contain the corresponding facial expression changes (\eg, ``much frowning" and ``gnashing with grimace'').
Compared to the DMD, our approach facilitates the corresponding attention module to focus on the intersection of meaningful signals between two modalities while mitigating the allocation bias of high attention scores.
We argue that this advantage benefits from aligning distribution domains and eliminating redundant information across modalities when learning modality-agnostic representations. Previous work~\cite{liang2021attention} has also evidenced that mitigating distributional differences can improve cross-modal correlations.
Overall, the above qualitative analyses of the PSA and HCA modules clearly show that the overall MEA can handle the temporal asynchrony challenge well.

\subsubsection{Visualization of Distinct Representations}
Observing the distribution of distinct representations is an effective measurement for determining whether the model is capable of dealing with the modality heterogeneity. As shown in Figure~\ref{vis_decoplued}, we visualize the modality-agnostic representations $\bm{h}^{a}_{m}$ and modality-exclusive representations $\bm{h}^{e}_{m}$ learned in the testing samples of the three datasets, where $m \in \{L, V, A\}$. 
When we remove the regularization constraints (\ie, $\alpha = 0, \beta =0 $), the decoupled representations are not well captured since most representations are unthinkingly distributed while occasionally overlapping. Most importantly, modality-agnostic representations $\bm{h}^{a}_{m}$ are not learned thoroughly.
Conversely, when we implement the necessary supervision (\ie, $\alpha \ne 0, \beta \ne 0 $), the distributions of $\bm{h}^{a}_{m}$ are effectively aggregated, where adversarial constraints judiciously bridge the domain gaps among the heterogeneous modalities. Meanwhile, each modality-exclusive subspace is well separated, where the disparity constraint punishes redundant latent representations.
The above observations clearly indicate that our approach addresses the modality heterogeneity challenge by capturing the commonality and diversity across multiple modalities.

\begin{figure*}[t]
  \centering
  \includegraphics[width=0.9\linewidth]{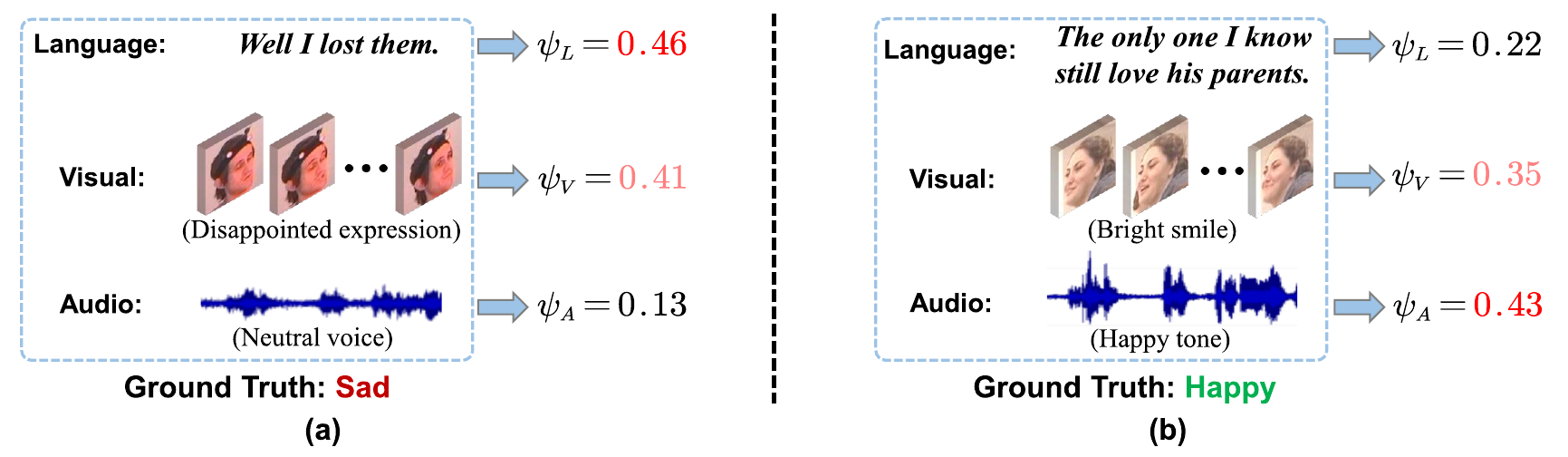}
  \caption{We show the attention weights $\psi_{m}$  from the weighted attention layer for the three modalities, where $ m\in \{ L,V,A\}$. Two randomly selected samples are from the IEMOCAP dataset.}
  \label{vis_wal}
\end{figure*}
\subsubsection{Effectiveness of Weighted Attention Layer}
Furthermore, we randomly select two examples on the IEMOCAP dataset to justify the role of our Weighted Attention Layer (WAL). From Figure~\ref{vis_wal}(a), the male speaker conveys sadness via the emotion-related word ``lost'' and the disappointed expression, while his voice is neutral. Therefore, WAL gives higher weights to language and visual modalities, \ie, $\psi_{L} = 0.46\,\&\,\psi_{V} = 0.41$. Meanwhile, the acoustic modality gets the lowest attention value (\ie, $\psi_{A} = 0.13$) since the neutral voice provides limited and ambiguous emotion clues.
In contrast, the female speaker in Figure~\ref{vis_wal}(b) has a happy tone and a bright smile, resulting in more significant attention values in the audio and visual modalities, \ie, $\psi_{A} = 0.43\,\&\,\psi_{V} = 0.35$.
These findings show that WAL can adaptively assign reasonable weights to different modalities based on their importance.

\subsection{Ablation Studies}
To systematically investigate the importance of all components and mechanisms in MEA, we perform thorough ablation studies on the MOSI and MOSEI datasets. We report representative $Acc_7$ and $F1$ scores while ignoring other metrics due to similar trends. Experimental results in Table~\ref{tab_abl} display the following observations.

\subsubsection{Necessity of Regularization}
We first explore the impact of regularization since it plays an essential role in disentangling different representations.

(i) As shown in the upper part of Table~\ref{tab_abl}, the disparity constraint $\mathcal{L}_{dis}$ provides beneficial gains to the model due to the performance drop when it is removed. Meanwhile, we provide an orthogonal-based separation constraint~\cite{bousmalis2016domain} $\mathcal{L}_{sep}$ to perform a comparison experiment.
Despite being adequate but inferior to $\mathcal{L}_{dis}$, which reveals that measuring independence between features can better distinguish distinct representations from heterogeneous modalities.

(ii) When the adversarial constraints $(\mathcal{L}_{agn}+ \mathcal{L}_{exc})$ are removed, the training procedure of our MEA does not involve the double-discriminator adversarial strategy.
In this case, the poor results clearly demonstrate the advantage of the adversarial paradigm in guiding the generation of decoupled representations.

(iii) Furthermore, we find that the degree $\omega^{a}_{m}$ from the importance discriminator fulfills the necessary effect as it promotes informative modality-agnostic representations.

\subsubsection{Importance of Representations}
Here, we observe the performance variation of the model by using only either representation in the feature aggregation phase. Both decreased results indicate the effectiveness of learning modality-exclusive and -agnostic representations.
This observation also justifies the feature decoupling insight of our MEA.
Another interesting finding is that the performance is worse without the modality-exclusive representations.
Although agnostic features are also important, exclusive representations reflect inherent properties and unique characteristics of each modality, providing complementary information in better improving affective semantics.

\subsubsection{Impact of PSA Module}
The PSA module is systematically investigated for its impact.

(i) Globally, when the PSA module is removed, the inevitable performance deterioration suggests that improved self-attention facilitates capturing intra-modal dynamics and reinforcing context-critical semantics.

(ii) Locally, we find that both the weighted attention layer and the prediction chain provide indispensable contributions to improve performance. The advantage of the weighted attention layer derives from dynamically assigning suitable importance to different modalities. The gains in the prediction chain are more significant, showing that the convolution-based attention pattern imparts valuable apriori knowledge to vanilla self-attention.

\begin{table*}[t]
\setlength{\tabcolsep}{15pt}
\centering
\caption{Ablation study results on the MOSI and MOSEI datasets. ``w/'' and ``w/o'' stand for the with and without, respectively. ``MRU'' means the Modality Reinforcement Unit in the HCA module. We report the $Acc_7$ and $F1$ scores for clarity.}
\resizebox{0.8\linewidth}{!}{%
\begin{tabular}{ccccc}
\toprule
\multicolumn{1}{c|}{\multirow{2}{*}{Components/Designs/Mechanisms}}     & \multicolumn{2}{c|}{MOSI}                          & \multicolumn{2}{c}{MOSEI}     \\ \cline{2-5} 
\multicolumn{1}{c|}{}                           & $Acc_7\uparrow$          & \multicolumn{1}{c|}{$F1\uparrow$}            & $Acc_7\uparrow$          & $F1\uparrow$            \\ \midrule
\multicolumn{1}{c|}{\textbf{MEA   (Full Model)}}               & \textbf{42.5} & \multicolumn{1}{c|}{\textbf{84.6}} & \textbf{54.8} & \textbf{85.1} \\ \hline
\multicolumn{5}{c}{Necessity   of Regularization}                                                                                    \\ \midrule
\multicolumn{1}{c|}{w/o  Disparity Constraint $\mathcal{L}_{dis}$}                 & 41.9          & \multicolumn{1}{c|}{83.5}          & 54.2          & 84.5          \\
\multicolumn{1}{c|}{w/  Separation Constraint $\mathcal{L}_{sep}$~\cite{bousmalis2016domain}}                  & 42.1          & \multicolumn{1}{c|}{84.2}          & 53.9          & 84.0          \\
\multicolumn{1}{c|}{w/o  Adversarial Constraints $\mathcal{L}_{agn}$ + $\mathcal{L}_{exc}$}           & 41.3          & \multicolumn{1}{c|}{82.9}          & 53.7          & 83.8          \\
\multicolumn{1}{c|}{w/o  Degree $\omega^{a}_{m}$}                    & 42.3          & \multicolumn{1}{c|}{84.5}          & 54.6          & 84.9          \\ \midrule
\multicolumn{5}{c}{Importance   of Representations}                                                                                  \\ \midrule
\multicolumn{1}{c|}{w/o   Modality-Exclusive}   & 39.4          & \multicolumn{1}{c|}{82.4}          & 51.7          & 81.9          \\
\multicolumn{1}{c|}{w/o   Modality-Agnostic}    & 40.6          & \multicolumn{1}{c|}{82.6}          & 52.4          & 82.8          \\ \midrule
\multicolumn{5}{c}{Impact of Predictive Self-Attention (PSA) Module}                                                                                           \\ \midrule
\multicolumn{1}{c|}{w/o   PSA Module}           & 38.7          & \multicolumn{1}{c|}{81.8}          & 51.2          & 81.5          \\
\multicolumn{1}{c|}{w/o   Prediction Chain}     & 41.6          & \multicolumn{1}{c|}{83.3}          & 53.5          & 83.6          \\
\multicolumn{1}{c|}{w/o   Weighted Attention Layer}                  & 41.8          & \multicolumn{1}{c|}{83.4}          & 54.1          & 84.4          \\ \midrule
\multicolumn{5}{c}{Impact of Hierarchical Cross-modal Attention (HCA) Module}                                                                                           \\ \midrule
\multicolumn{1}{c|}{w/o   HCA Module}           & 39.5          & \multicolumn{1}{c|}{82.6}          & 51.9          & 82.2          \\
\multicolumn{1}{c|}{w/o   MRU (Mixed-grained)}  & 41.5          & \multicolumn{1}{c|}{83.3}          & 53.6          & 83.8          \\
\multicolumn{1}{c|}{w/o   MRU (Coarse-grained)} & 42.3          & \multicolumn{1}{c|}{84.4}          & 54.5          & 84.8          \\
\multicolumn{1}{c|}{w/o   MRU (Fine-grained)}   & 41.2          & \multicolumn{1}{c|}{82.8}          & 53.3          & 83.4          \\ \midrule
\multicolumn{5}{c}{Effectiveness   of Fusion Mechanisms}                                                                              \\ \midrule
\multicolumn{1}{c|}{w/o   Decoupled Graph Fusion  Mechanism}                  & 41.6          & \multicolumn{1}{c|}{83.4}          & 53.5          & 83.6          \\
\multicolumn{1}{c|}{w/   Feature Addition}              & 40.1          & \multicolumn{1}{c|}{82.5}          & 52.7          & 82.9          \\
\multicolumn{1}{c|}{w/  Feature Multiplication~\cite{yang2022emotion}}        & 41.8          & \multicolumn{1}{c|}{83.5}          & 53.3          & 83.4          \\ \bottomrule
\end{tabular}
}
\label{tab_abl}
\end{table*}

\begin{figure}[t]
  \centering
  \includegraphics[width=\linewidth]{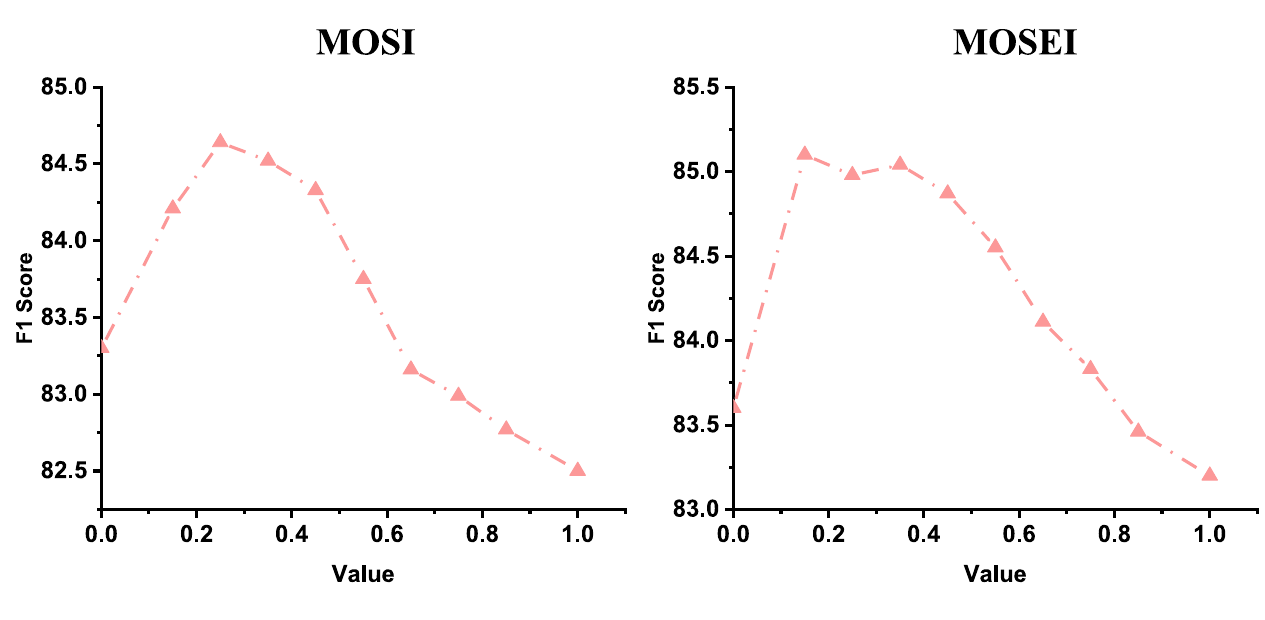}
  \caption{We show the effect of the coefficient $\mu$ on model performance over the two datasets by controlling for different values. The results are obtained by changing the values of the corresponding hyper-parameter, while keeping the values of the other hyper-parameters fixed.}
  \label{sen_u}
\end{figure}

\begin{figure}[t]
  \centering
  \includegraphics[width=\linewidth]{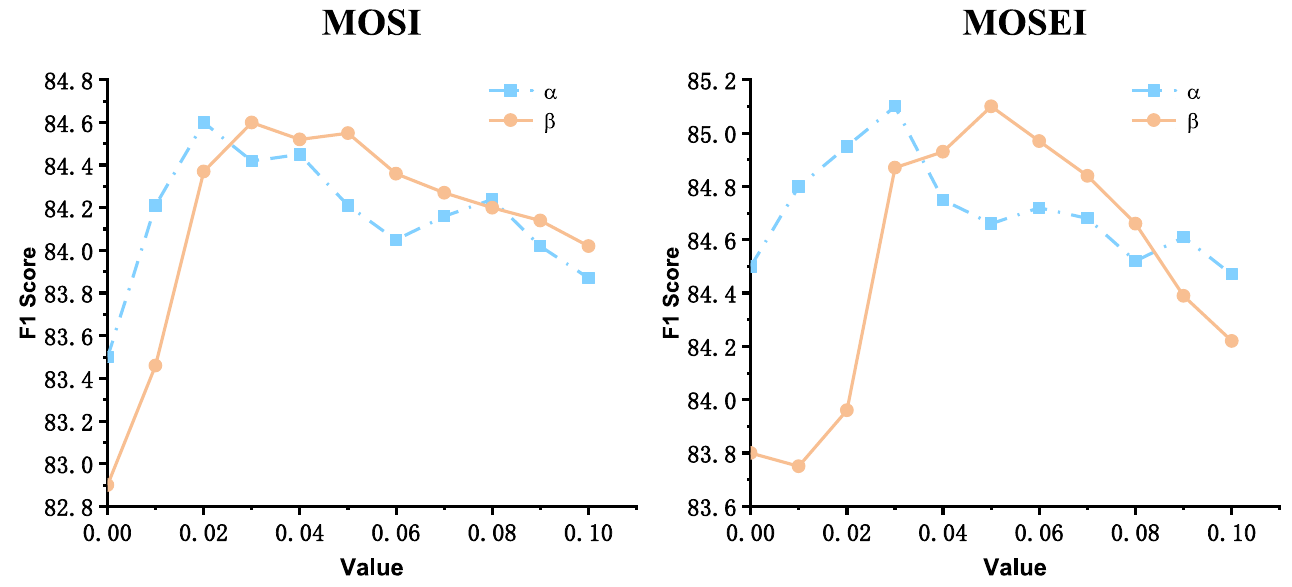}
  \caption{We show the effect of the parameters $\alpha\&\beta$ on model performance over the two datasets by controlling for different values. The results are obtained by changing the values of the corresponding hyper-parameter, while keeping the values of the other hyper-parameters fixed.}
  \label{sen_ab}
\end{figure}

\subsubsection{Impact of HCA Module}
Immediately, we evaluate the impact of the HCA module.

(i) First, the complete HCA module offers effective performance gains on both datasets through comprehensive cross-modal attention interactions. This finding also confirms that simple attention is insufficient, as it fails to capture meaningful clues in cross-modal element correlations. 

(ii) For the HCA module internals, the mixed-grained, coarse-grained, and fine-grained modality reinforcement units (MRUs) are separately removed to explore the gains from the hierarchical structure.
The experimental results demonstrate the usefulness of performing cross-modal interactions at different granularities, where the element correlations over asynchronous sequences are fully explored.

(iii) Moreover, we perceive that the model seems to be more responsive to the fine-grained and mixed-grained MRUs.
This phenomenon inspires us to focus more on pairwise interactions between individual modalities and integrated interactions among multiple modalities in cross-modal designs.

\subsubsection{Effectiveness of Fusion Mechanisms}
Ultimately, we assess the effect of different fusion mechanisms. 
The feature fusion is performed with the simple feature concatenation when our decoupled graph fusion mechanism is removed. The significant gain deterioration demonstrates the effectiveness of our fusion strategy.
As a comparative alternative, feature addition fails to address the multi-feature fusion challenge as it potentially couples up the refined decoupled representations.
Additionally, our fusion mechanism remains competitive compared to the advanced multiplicative fusion~\cite{yang2022emotion}.

\subsection{Sensitivity Analysis}

\subsubsection{Analysis of the Coefficient $\mu$}
In Figure~\ref{sen_u}, we analyze the coefficient $\mu$ used in the Predictive Self-Attention (PSA) module to balance the prediction chain and the dot-product attention branch on the two datasets. Specifically, the values are varied from 0 to 1 to observe the trend in the $F1$ score of the model.  We find that as the values increase, the performances rise first and then drop significantly. Our model achieves the best performance when the coefficient $\mu$ of the MOSI and MOSEI datasets are set to 0.25 and 0.15, which aligns with the values adopted in the experiments. These observations suggest that it is essential to introduce predictive attention maps with the appropriate balance. Also, the model achieves the worst performance when 
$\mu$ is set to 1. This fact shows the dominant role that dot-product attention still plays in the PSA module.

\subsubsection{Analysis of the Trade-off Parameters $\alpha$ and $\beta$}

The trade-off parameter analysis on constraints is performed in Figure~\ref{sen_ab}.
The tested hyper-parameters include the parameter $\alpha$ for the disparity constraint $\mathcal{L}_{dis}$ and the parameter $\beta$ for the adversarial constraints ($\mathcal{L}_{agn}+ \mathcal{L}_{exc}$).
For both datasets, the $F1$ scores first increase and then start to decrease, where the results for the parameter $\alpha$ show a slightly oscillating trend. The proposed MEA achieves the best performance when the trade-off parameters $\alpha$ and $\beta$ of the MOSI and MOSEI datasets are set to $\{2e^{-2}, 3e^{-2} \}$ and $\{3e^{-2}, 5e^{-2}\}$, respectively. These appropriate balances are consistent with the values adopted in the experiments.
Overall, we believe that seeking a reasonable trade-off in constraints is critical to learning effective decoupled representations.

\begin{figure}[t]
  \centering
  \includegraphics[width=0.8\linewidth]{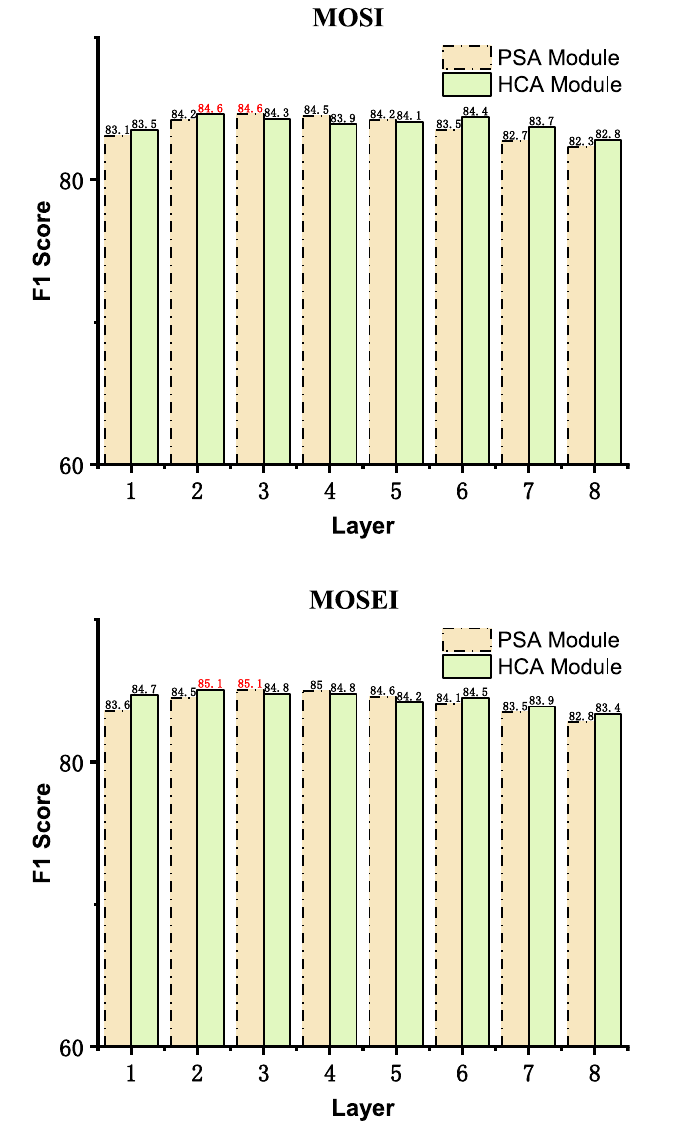}
  \caption{We show the sensitivity analysis of the number of the PSA and HCA layers on the two datasets. The results are obtained by changing the values of the corresponding hyper-parameter, while keeping the values of the other hyper-parameters fixed.}
  \label{layer}
\end{figure}

\subsubsection{Analysis of the Number of PSA and HCA Layers}
Figure~\ref{layer} shows the effect of varying the number of layers in the PSA and HCA modules on model performance.

(i) For the PSA module, the performances decrease significantly when the number of layers exceeds 3 on both datasets.
For the HCA module, the performances gradually increase and stabilize as the number of layers rises on both datasets.

(ii) A noteworthy observation is that the deep module stacking leads to unavoidable gain degradation, suggesting that overly complex structures would bring performance bottlenecks.
Compared to the previous SOTA models~\cite{yang2022disentangled,tsai2019multimodal,lv2021progressive,liang2021attention,li2023decoupled,yang2023target} with extensive parameters and module stacking, MEA achieves superior performance with fewer parameters (\ie, a smaller number of layers).

\section{Conclusion and Discussion}
\label{sec6}
In this paper, we present MEA, an asynchronous multimodal fusion approach to refine and reinforce multimodal representations progressively. On the one hand, MEA adequately explores intra-modal dynamics and cross-modal interactions to alleviate the temporal asynchronicity dilemma based on improved attention patterns. The customized attention modules capture reliable element correlations and context dependencies.
On the other hand, MEA addresses the modality heterogeneity challenge based on feature disentanglement thinking for learning modality-exclusive and modality-agnostic representations. The tailored decoupled components learn valuable semantics commonalities among modalities while emphasizing the intrinsic characteristics of each modality. Extensive experiments prove the necessity of the proposed components.

\textbf{Discussion of Broad Impacts.}
(i)
This study can overcome the noisy interference and redundant information in heterogeneous modality sequences to improve the accuracy in diverse real-world applications such as video-based anomaly detection, sentiment analysis, and action expectation recognition. 
(ii)
The proposed feature decoupling concept facilitates a suitable generalization of our approach to out-of-distribution scenarios to tackle complicated information devices through flexible attention modules.
(iii) MEA can be readily extended to other multimodal fusion tasks to provide pragmatic perception outcomes. For instance, our algorithm can enhance therapeutic effects by fusing X-ray and MRI information in disease diagnosis.
In intelligent transportation, our algorithm allows fine-grained fusion between camera and LiDAR data, facilitating comprehensive scene perception to reduce traffic accidents.

\textbf{Discussion of Limitation and Future Work.}
Our approach relies on the setting of complete input modalities. When the model is applied to scenarios with missing modalities, the decoupled procedure from different modalities is potentially limited and weakened, resulting in sub-optimal improvements.
Moreover, when there is data distribution bias in the training samples, our approach may suffer from bias interference to induce feature entanglement in the decoupled modality subspaces, weakening the robustness of representation learning.
In the future, we plan to equip MEA with
modality reconstruction and causal inference techniques to cope with potential
modality missingness and data bias in realistic applications.

\bibliographystyle{IEEEtran}
\bibliography{tcsvt}

\begin{thebibliography}{10}
\providecommand{\url}[1]{#1}
\csname url@samestyle\endcsname
\providecommand{\newblock}{\relax}
\providecommand{\bibinfo}[2]{#2}
\providecommand{\BIBentrySTDinterwordspacing}{\spaceskip=0pt\relax}
\providecommand{\BIBentryALTinterwordstretchfactor}{4}
\providecommand{\BIBentryALTinterwordspacing}{\spaceskip=\fontdimen2\font plus
\BIBentryALTinterwordstretchfactor\fontdimen3\font minus \fontdimen4\font\relax}
\providecommand{\BIBforeignlanguage}[2]{{%
\expandafter\ifx\csname l@#1\endcsname\relax
\typeout{** WARNING: IEEEtran.bst: No hyphenation pattern has been}%
\typeout{** loaded for the language `#1'. Using the pattern for}%
\typeout{** the default language instead.}%
\else
\language=\csname l@#1\endcsname
\fi
#2}}
\providecommand{\BIBdecl}{\relax}
\BIBdecl

\bibitem{zhang2021multimodal}
L.~Zhang, J.~Shen, J.~Zhang, J.~Xu, Z.~Li, Y.~Yao, and L.~Yu, ``Multimodal marketing intent analysis for effective targeted advertising,'' \emph{IEEE Trans. Multimedia}, vol.~24, pp. 1830--1843, 2021.

\bibitem{marivani2022designing}
I.~Marivani, E.~Tsiligianni, B.~Cornelis, and N.~Deligiannis, ``Designing cnns for multimodal image restoration and fusion via unfolding the method of multipliers,'' \emph{IEEE Trans. Circuits Syst. Video Technol.}, vol.~32, no.~9, pp. 5830--5845, 2022.

\bibitem{zhou2019understanding}
S.~Zhou, J.~Jia, Y.~Yin, X.~Li, Y.~Yao, Y.~Zhang, Z.~Ye, K.~Lei, Y.~Huang, and J.~Shen, ``Understanding the teaching styles by an attention based multi-task cross-media dimensional modeling,'' in \emph{Proc. ACM Int. Conf. Multimedia}, 2019, pp. 1322--1330.

\bibitem{feng2022temporal}
Z.~Feng, Z.~Zeng, C.~Guo, and Z.~Li, ``Temporal multimodal graph transformer with global-local alignment for video-text retrieval,'' \emph{IEEE Trans. Circuits Syst. Video Technol.}, vol.~33, no.~3, pp. 1438--1453, 2022.

\bibitem{shen2021bbas}
J.~Shen and N.~Robertson, ``Bbas: Towards large scale effective ensemble adversarial attacks against deep neural network learning,'' \emph{Inform. Sciences}, vol. 569, pp. 469--478, 2021.

\bibitem{yang2022disentangled}
D.~Yang, S.~Huang, H.~Kuang, Y.~Du, and L.~Zhang, ``Disentangled representation learning for multimodal emotion recognition,'' in \emph{Proc. ACM Int. Conf. Multimedia}, 2022, p. 1642–1651.

\bibitem{yang2022learning}
D.~Yang, H.~Kuang, S.~Huang, and L.~Zhang, ``Learning modality-specific and -agnostic representations for asynchronous multimodal language sequences,'' in \emph{Proc. ACM Int. Conf. Multimedia}, 2022, p. 1708–1717.

\bibitem{lei2023text}
Y.~Lei, D.~Yang, M.~Li, S.~Wang, J.~Chen, and L.~Zhang, ``Text-oriented modality reinforcement network for multimodal sentiment analysis from unaligned multimodal sequences,'' \emph{arXiv preprint arXiv:2307.13205}, 2023.

\bibitem{zadeh2016multimodal}
A.~Zadeh, R.~Zellers, E.~Pincus, and L.-P. Morency, ``Multimodal sentiment intensity analysis in videos: Facial gestures and verbal messages,'' \emph{IEEE Intell. Syst.}, vol.~31, no.~6, pp. 82--88, 2016.

\bibitem{yang2023target}
D.~Yang, Y.~Liu, C.~Huang, M.~Li, X.~Zhao, Y.~Wang, K.~Yang, Y.~Wang, P.~Zhai, and L.~Zhang, ``Target and source modality co-reinforcement for emotion understanding from asynchronous multimodal sequences,'' \emph{Knowl.-Based Syst.}, vol. 265, p. 110370, 2023.

\bibitem{yang2022contextual}
D.~Yang, S.~Huang, Y.~Liu, and L.~Zhang, ``Contextual and cross-modal interaction for multi-modal speech emotion recognition,'' \emph{IEEE Signal Process. Lett.}, vol.~29, pp. 2093--2097, 2022.

\bibitem{tsai2019multimodal}
Y.-H.~H. Tsai, S.~Bai, P.~P. Liang, J.~Z. Kolter, L.-P. Morency, and R.~Salakhutdinov, ``Multimodal transformer for unaligned multimodal language sequences,'' in \emph{Proc. Conf. Annu. Meet. Assoc. Comput. Linguist.}, vol. 2019, 2019, p. 6558.

\bibitem{wang2019words}
Y.~Wang, Y.~Shen, Z.~Liu, P.~P. Liang, A.~Zadeh, and L.-P. Morency, ``Words can shift: Dynamically adjusting word representations using nonverbal behaviors,'' in \emph{Proc. AAAI Conf. Artif. Intell.}, 2019, pp. 7216--7223.

\bibitem{pham2019found}
H.~Pham, P.~P. Liang, T.~Manzini, L.-P. Morency, and B.~P{\'o}czos, ``Found in translation: Learning robust joint representations by cyclic translations between modalities,'' in \emph{Proc. AAAI Conf. Artif. Intell.}, 2019, pp. 6892--6899.

\bibitem{tsai2018learning}
Y.-H.~H. Tsai, P.~P. Liang, A.~Zadeh, L.-P. Morency, and R.~Salakhutdinov, ``Learning factorized multimodal representations,'' \emph{arXiv preprint arXiv:1806.06176}, 2018.

\bibitem{wu2021text}
Y.~Wu, Z.~Lin, Y.~Zhao, B.~Qin, and L.-N. Zhu, ``A text-centered shared-private framework via cross-modal prediction for multimodal sentiment analysis,'' in \emph{Assoc. Comput. Linguistics: ACL-IJCNLP 2021}, 2021, pp. 4730--4738.

\bibitem{liang2018multimodal}
P.~P. Liang, Z.~Liu, A.~Zadeh, and L.-P. Morency, ``Multimodal language analysis with recurrent multistage fusion,'' \emph{arXiv preprint arXiv:1808.03920}, 2018.

\bibitem{rahman2020integrating}
W.~Rahman, M.~K. Hasan, S.~Lee, A.~Zadeh, C.~Mao, L.-P. Morency, and E.~Hoque, ``Integrating multimodal information in large pretrained transformers,'' in \emph{Proc. Conf. Annu. Meet. Assoc. Comput. Linguist.}, vol. 2020, 2020, p. 2359.

\bibitem{wu2023denoising}
S.~Wu, D.~Dai, Z.~Qin, T.~Liu, B.~Lin, Y.~Cao, and Z.~Sui, ``Denoising bottleneck with mutual information maximization for video multimodal fusion,'' \emph{arXiv preprint arXiv:2305.14652}, 2023.

\bibitem{wang2023cross}
Y.~Wang, Y.~Li, P.~Bell, and C.~Lai, ``Cross-attention is not enough: Incongruity-aware multimodal sentiment analysis and emotion recognition,'' \emph{arXiv preprint arXiv:2305.13583}, 2023.

\bibitem{sahay2020low}
S.~Sahay, E.~Okur, S.~H. Kumar, and L.~Nachman, ``Low rank fusion based transformers for multimodal sequences,'' \emph{arXiv preprint arXiv:2007.02038}, 2020.

\bibitem{lv2021progressive}
F.~Lv, X.~Chen, Y.~Huang, L.~Duan, and G.~Lin, ``Progressive modality reinforcement for human multimodal emotion recognition from unaligned multimodal sequences,'' in \emph{Proc. IEEE/CVF Conf. Comput. Vis. Pattern Recognit.}, 2021, pp. 2554--2562.

\bibitem{liang2021attention}
T.~Liang, G.~Lin, L.~Feng, Y.~Zhang, and F.~Lv, ``Attention is not enough: Mitigating the distribution discrepancy in asynchronous multimodal sequence fusion,'' in \emph{Proc. IEEE Int. Conf. Comput. Vis.}, 2021, pp. 8148--8156.

\bibitem{vaswani2017attention}
A.~Vaswani, N.~Shazeer, N.~Parmar, J.~Uszkoreit, L.~Jones, A.~N. Gomez, {\L}.~Kaiser, and I.~Polosukhin, ``Attention is all you need,'' in \emph{Adv. Neural Inf. Process. Syst.}, 2017, pp. 5998--6008.

\bibitem{Hazarika2020mm}
H.~Devamanyu, Z.~Roger, and P.~Soujanya, ``Misa: Modality-invariant and-specific representations for multimodal sentiment analysis.'' in \emph{Proc. ACM Int. Conf. Multimedia}, 2020, p. 1122–1131.

\bibitem{li2023decoupled}
Y.~Li, Y.~Wang, and Z.~Cui, ``Decoupled multimodal distilling for emotion recognition,'' in \emph{Proc. IEEE/CVF Conf. Comput. Vis. Pattern Recognit.}, 2023, pp. 6631--6640.

\bibitem{liu2023stochastic}
Y.~Liu, D.~Yang, G.~Fang, Y.~Wang, D.~Wei, M.~Zhao, K.~Cheng, J.~Liu, and L.~Song, ``Stochastic video normality network for abnormal event detection in surveillance videos,'' \emph{Knowl-based Syst.}, vol. 280, p. 110986, 2023.

\bibitem{liu2024memory}
Y.~Liu, B.~Ju, D.~Yang, L.~Peng, D.~Li, P.~Sun, C.~Li, H.~Yang, J.~Liu, and L.~Song, ``Memory-enhanced spatial-temporal encoding framework for industrial anomaly detection system,'' \emph{Expert Syst. Appl.}, vol. 250, p. 123718, 2024.

\bibitem{zhang2023transformer}
X.~Zhang, M.~Li, S.~Lin, H.~Xu, and G.~Xiao, ``Transformer-based multimodal emotional perception for dynamic facial expression recognition in the wild,'' \emph{IEEE Trans. Circuits Syst. Video Technol.}, vol.~34, no.~5, pp. 3192--3203, 2024.

\bibitem{tang2023tldw}
P.~Tang, K.~Hu, L.~Zhang, J.~Luo, and Z.~Wang, ``Tldw: Extreme multimodal summarisation of news videos,'' \emph{IEEE Trans. Circuits Syst. Video Technol.}, vol.~34, no.~3, pp. 1469--1480, 2024.

\bibitem{yang2023aide}
D.~Yang, S.~Huang, Z.~Xu, Z.~Li, S.~Wang, M.~Li, Y.~Wang, Y.~Liu, K.~Yang, Z.~Chen, Y.~Wang, J.~Liu, P.~Zhang, P.~Zhai, and L.~Zhang, ``Aide: A vision-driven multi-view, multi-modal, multi-tasking dataset for assistive driving perception,'' in \emph{Proc. IEEE Int. Conf. Comput. Vis.}, October 2023, pp. 20\,459--20\,470.

\bibitem{he2023multimodal}
L.~He, Z.~Wang, L.~Wang, and F.~Li, ``Multimodal mutual attention-based sentiment analysis framework adapted to complicated contexts,'' \emph{IEEE Trans. Circuits Syst. Video Technol.}, vol.~33, no.~12, pp. 7131--7143, 2023.

\bibitem{fu2022drake}
Z.~Fu, C.~Zheng, J.~Feng, Y.~Cai, X.-Y. Wei, Y.~Wang, and Q.~Li, ``Drake: Deep pair-wise relation alignment for knowledge-enhanced multimodal scene graph generation in social media posts,'' \emph{IEEE Trans. Circuits Syst. Video Technol.}, vol.~33, no.~7, pp. 3199--3213, 2023.

\bibitem{guan2023egocentric}
W.~Guan, X.~Song, K.~Wang, H.~Wen, H.~Ni, Y.~Wang, and X.~Chang, ``Egocentric early action prediction via multimodal transformer-based dual action prediction,'' \emph{IEEE Trans. Circuits Syst. Video Technol.}, vol.~33, no.~9, pp. 4472--4483, 2023.

\bibitem{li2023towards}
M.~Li, D.~Yang, and L.~Zhang, ``Towards robust multimodal sentiment analysis under uncertain signal missing,'' \emph{IEEE Signal Process. Lett.}, vol.~30, pp. 1497--1501, 2023.

\bibitem{li2024unified}
M.~Li, D.~Yang, Y.~Lei, S.~Wang, S.~Wang, L.~Su, K.~Yang, Y.~Wang, M.~Sun, and L.~Zhang, ``A unified self-distillation framework for multimodal sentiment analysis with uncertain missing modalities,'' in \emph{Proc. AAAI Conf. Artif. Intell.}, vol.~38, 2024, pp. 10\,074--10\,082.

\bibitem{yang2024SuCi}
D.~Yang, D.~Xiao, K.~Li, Y.~Wang, Z.~Chen, J.~Wei, and L.~Zhang, ``Towards multimodal human intention understanding debiasing via subject-deconfounding,'' \emph{arXiv preprint arXiv:2403.05025}, 2024.

\bibitem{liu2024generalized}
Y.~Liu, D.~Yang, Y.~Wang, J.~Liu, J.~Liu, A.~Boukerche, P.~Sun, and L.~Song, ``Generalized video anomaly event detection: Systematic taxonomy and comparison of deep models,'' \emph{ACM Comput. Surv.}, vol.~56, no.~7, pp. 1--38, 2024.

\bibitem{MRG3Net10419057}
Y.~Wang, S.~Yan, W.~Song, A.~Liotta, J.~Liu, D.~Yang, S.~Gao, and W.~Zhang, ``Mgr3net: Multigranularity region relation representation network for facial expression recognition in affective robots,'' \emph{IEEE Trans. Ind. Inf.}, vol.~20, no.~5, pp. 7216--7226, 2024.

\bibitem{yang2024TSIF}
D.~Yang, H.~Kuang, K.~Yang, M.~Li, and L.~Zhang, ``Towards asynchronous multimodal signal interaction and fusion via tailored transformers,'' \emph{IEEE Signal Process. Lett.}, 2024.

\bibitem{huang2021emotion}
Y.~Huang, H.~Wen, L.~Qing, R.~Jin, and L.~Xiao, ``Emotion recognition based on body and context fusion in the wild,'' in \emph{Proc. IEEE Int. Conf. Comput. Vis. Workshop}, 2021, pp. 3609--3617.

\bibitem{duan2021multi}
W.~Duan, L.~Zhang, J.~Colman, G.~Gulli, and X.~Ye, ``Multi-modal brain segmentation using hyper-fused convolutional neural network,'' in \emph{Int. Workshop Mach. Learn. in Clin. Neuroimaging}.\hskip 1em plus 0.5em minus 0.4em\relax Springer, 2021, pp. 82--91.

\bibitem{yang2024MCIS}
D.~Yang, M.~Li, D.~Xiao, Y.~Liu, K.~Yang, Z.~Chen, Y.~Wang, P.~Zhai, K.~Li, and L.~Zhang, ``Towards multimodal sentiment analysis debiasing via bias purification,'' \emph{arXiv preprint arXiv:2403.05023}, 2024.

\bibitem{yang2024towards}
D.~Yang, H.~Kuang, K.~Yang, M.~Li, and L.~Zhang, ``Towards asynchronous multimodal signal interaction and fusion via tailored transformers,'' \emph{IEEE Signal Process. Lett.}, vol.~30, pp. 1550--1554, 2024.

\bibitem{medsker2001recurrent}
L.~R. Medsker and L.~Jain, ``Recurrent neural networks,'' \emph{Des. Applic.}, vol.~5, pp. 64--67, 2001.

\bibitem{hochreiter1997long}
S.~Hochreiter and J.~Schmidhuber, ``Long short-term memory,'' \emph{Neural Comput.}, vol.~9, no.~8, pp. 1735--1780, 1997.

\bibitem{liu2018efficient}
Z.~Liu, Y.~Shen, V.~B. Lakshminarasimhan, P.~P. Liang, A.~Zadeh, and L.-P. Morency, ``Efficient low-rank multimodal fusion with modality-specific factors,'' \emph{arXiv preprint arXiv:1806.00064}, 2018.

\bibitem{yang2024pediatricsgpt}
D.~Yang, J.~Wei, D.~Xiao, S.~Wang, T.~Wu, G.~Li, M.~Li, S.~Wang, J.~Chen, Y.~Jiang \emph{et~al.}, ``Pediatricsgpt: Large language models as chinese medical assistants for pediatric applications,'' \emph{arXiv preprint arXiv:2405.19266}, 2024.

\bibitem{yang2024robust}
D.~Yang, K.~Yang, M.~Li, S.~Wang, S.~Wang, and L.~Zhang, ``Robust emotion recognition in context debiasing,'' \emph{arXiv preprint arXiv:2403.05963}, 2024.

\bibitem{yang2023how2comm}
D.~Yang, K.~Yang, Y.~Wang, J.~Liu, Z.~Xu, R.~Yin, P.~Zhai, and L.~Zhang, ``How2comm: Communication-efficient and collaboration-pragmatic multi-agent perception,'' in \emph{Adv. Neural Inf. Process. Syst.}, 2023.

\bibitem{du2021learning}
Y.~Du, D.~Yang, P.~Zhai, M.~Li, and L.~Zhang, ``Learning associative representation for facial expression recognition,'' in \emph{Proc. Int. Conf. Image Process.}, 2021, pp. 889--893.

\bibitem{yang2023context}
D.~Yang, Z.~Chen, Y.~Wang, S.~Wang, M.~Li, S.~Liu, X.~Zhao, S.~Huang, Z.~Dong, P.~Zhai, and L.~Zhang, ``Context de-confounded emotion recognition,'' in \emph{Proc. IEEE/CVF Conf. Comput. Vis. Pattern Recognit.}, June 2023, pp. 19\,005--19\,015.

\bibitem{wang2022spacenet}
S.~Wang, S.~Wang, B.~Jiao, D.~Yang, L.~Su, P.~Zhai, C.~Chen, and L.~Zhang, ``Ca-spacenet: Counterfactual analysis for 6d pose estimation in space,'' in \emph{IEEE/RSJ Int. Conf. on Intell. Robot. Syst.}\hskip 1em plus 0.5em minus 0.4em\relax IEEE, 2022, pp. 10\,627--10\,634.

\bibitem{zhang2019multimodal}
S.-F. Zhang, J.-H. Zhai, B.-J. Xie, Y.~Zhan, and X.~Wang, ``Multimodal representation learning: advances, trends and challenges,'' in \emph{Int. Conf. Mach. Learn. Cybern.}, 2019, pp. 1--6.

\bibitem{yu2021learning}
W.~Yu, H.~Xu, Z.~Yuan, and J.~Wu, ``Learning modality-specific representations with self-supervised multi-task learning for multimodal sentiment analysis,'' in \emph{Proc. AAAI Conf. Artif. Intell.}, 2021, pp. 10\,790--10\,797.

\bibitem{zhang2022tailor}
Y.~Zhang, M.~Chen, J.~Shen, and C.~Wang, ``Tailor versatile multi-modal learning for multi-label emotion recognition,'' in \emph{Proc. AAAI Conf. Artif. Intell.}, 2022, pp. 9100--9108.

\bibitem{park2016image}
G.~Park and W.~Im, ``Image-text multi-modal representation learning by adversarial backpropagation,'' \emph{arXiv preprint arXiv:1612.08354}, 2016.

\bibitem{2020Learning}
Z.~Sun, P.~Sarma, W.~Sethares, and Y.~Liang, ``Learning relationships between text, audio, and video via deep canonical correlation for multimodal language analysis,'' \emph{Proc. AAAI Conf. Artif. Intell.}, pp. 8992--8999, 2020.

\bibitem{bousmalis2016domain}
K.~Bousmalis, G.~Trigeorgis, N.~Silberman, D.~Krishnan, and D.~Erhan, ``Domain separation networks,'' \emph{Adv. Neural Inf. Process. Syst.}, vol.~29, 2016.

\bibitem{chen2024can}
J.~Chen, Y.~Jiang, D.~Yang, M.~Li, J.~Wei, Z.~Qian, and L.~Zhang, ``Can llms' tuning methods work in medical multimodal domain?'' \emph{arXiv preprint arXiv:2403.06407}, 2024.

\bibitem{chen2024efficiency}
J.~Chen, D.~Yang, Y.~Jiang, M.~Li, J.~Wei, X.~Hou, and L.~Zhang, ``Efficiency in focus: Layernorm as a catalyst for fine-tuning medical visual language pre-trained models,'' \emph{arXiv preprint arXiv:2404.16385}, 2024.

\bibitem{chen2024detecting}
J.~Chen, D.~Yang, T.~Wu, Y.~Jiang, X.~Hou, M.~Li, S.~Wang, D.~Xiao, K.~Li, and L.~Zhang, ``Detecting and evaluating medical hallucinations in large vision language models,'' \emph{arXiv preprint arXiv:2406.10185}, 2024.

\bibitem{jiang2024medthink}
Y.~Jiang, J.~Chen, D.~Yang, M.~Li, S.~Wang, T.~Wu, K.~Li, and L.~Zhang, ``Medthink: Inducing medical large-scale visual language models to hallucinate less by thinking more,'' \emph{arXiv preprint arXiv:2406.11451}, 2024.

\bibitem{jain2019attention}
S.~Jain and B.~C. Wallace, ``Attention is not explanation,'' \emph{arXiv preprint arXiv:1902.10186}, 2019.

\bibitem{wang2020predictive}
Y.~Wang, Y.~Yang, J.~Bai, M.~Zhang, J.~Bai, J.~Yu, C.~Zhang, and Y.~Tong, ``Predictive attention transformer: Improving transformer with attention map prediction,'' 2020.

\bibitem{hendrycks2016gaussian}
D.~Hendrycks and K.~Gimpel, ``Gaussian error linear units (gelus),'' \emph{arXiv preprint arXiv:1606.08415}, 2016.

\bibitem{song2007supervised}
L.~Song, A.~Smola, A.~Gretton, K.~M. Borgwardt, and J.~Bedo, ``Supervised feature selection via dependence estimation,'' in \emph{Proc. Int. Conf. Mach. Learn.}, 2007, pp. 823--830.

\bibitem{ganin2015unsupervised}
Y.~Ganin and V.~Lempitsky, ``Unsupervised domain adaptation by backpropagation,'' in \emph{Proc. Int. Conf. Mach. Learn.}, 2015, pp. 1180--1189.

\bibitem{zadeh2018multimodal}
A.~Zadeh and P.~Pu, ``Multimodal language analysis in the wild: Cmu-mosei dataset and interpretable dynamic fusion graph,'' in \emph{Proc. Conf. Annu. Meet. Assoc. Comput. Linguist.}, 2018, pp. 2236--2246.

\bibitem{busso2008iemocap}
C.~Busso, M.~Bulut, C.-C. Lee, A.~Kazemzadeh, E.~Mower, S.~Kim, J.~N. Chang, S.~Lee, and S.~S. Narayanan, ``Iemocap: Interactive emotional dyadic motion capture database,'' \emph{Lang. Resour. Eval.}, vol.~42, no.~4, pp. 335--359, 2008.

\bibitem{han2021improving}
W.~Han, H.~Chen, and S.~Poria, ``Improving multimodal fusion with hierarchical mutual information maximization for multimodal sentiment analysis,'' \emph{arXiv preprint arXiv:2109.00412}, 2021.

\bibitem{pennington2014glove}
J.~Pennington, R.~Socher, and C.~D. Manning, ``Glove: Global vectors for word representation,'' in \emph{Proc. Conf. Empir. Methods Nat. Lang. Process.}, 2014, pp. 1532--1543.

\bibitem{baltruvsaitis2016openface}
T.~Baltru{\v{s}}aitis, P.~Robinson, and L.-P. Morency, ``Openface: an open source facial behavior analysis toolkit,'' in \emph{Proc. IEEE Winter Conf. Appl. Comput. Vis.}, 2016, pp. 1--10.

\bibitem{degottex2014covarep}
G.~Degottex, J.~Kane, T.~Drugman, T.~Raitio, and S.~Scherer, ``Covarep—a collaborative voice analysis repository for speech technologies,'' in \emph{Proc. IEEE Int. Conf. Acoust., Speech Signal Process.}, 2014, pp. 960--964.

\bibitem{kingma2014adam}
D.~P. Kingma and J.~Ba, ``Adam: A method for stochastic optimization,'' \emph{arXiv preprint arXiv:1412.6980}, 2014.

\bibitem{graves2006connectionist}
A.~Graves, S.~Fern{\'a}ndez, F.~Gomez, and J.~Schmidhuber, ``Connectionist temporal classification: labelling unsegmented sequence data with recurrent neural networks,'' in \emph{Proc. Int. Conf. Mach. Learn.}, 2006, pp. 369--376.

\bibitem{yang2022emotion}
D.~Yang, S.~Huang, S.~Wang, Y.~Liu, P.~Zhai, L.~Su, M.~Li, and L.~Zhang, ``Emotion recognition for multiple context awareness,'' in \emph{Proc. Eur. Conf. Comput. Vis.}, vol. 13697, 2022, pp. 144--162.

\end{thebibliography}

\begin{IEEEbiography}[{\includegraphics[width=1in,height=1.25in,clip,keepaspectratio]{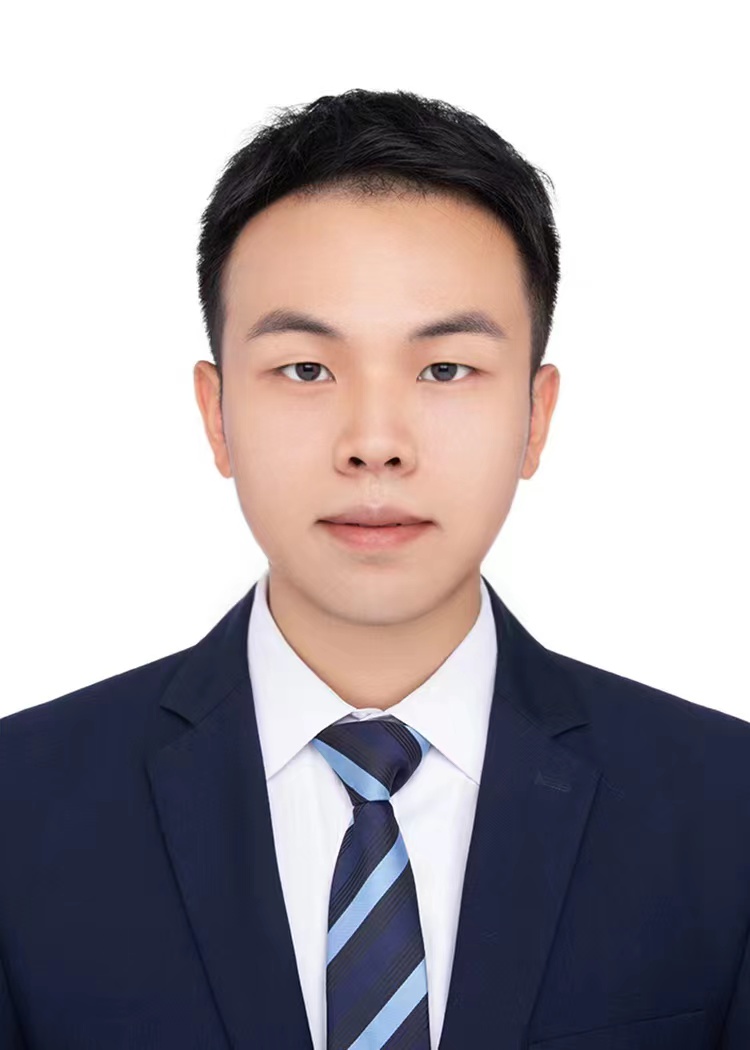}}]{Dingkang Yang} received the B.E. degree in Communication Engineering from the joint training of Yunnan University and the Chinese People's Armed Police (PAP), Kunming, China, in 2020. He is currently pursuing the Ph.D. degree at the Academy for Engineering and Technology, Fudan University, Shanghai, China. His research interests include multimodal learning, affective computing, large language models, and causal inference.
He has published over 35 papers in renowned international conferences, appearing in prestigious international academic conferences such as NeurIPS, CVPR, ICCV, ECCV, and AAAI.
\end{IEEEbiography}
\vspace{-1cm}

\begin{IEEEbiography}[{\includegraphics[width=1in,height=1.25in,clip,keepaspectratio]{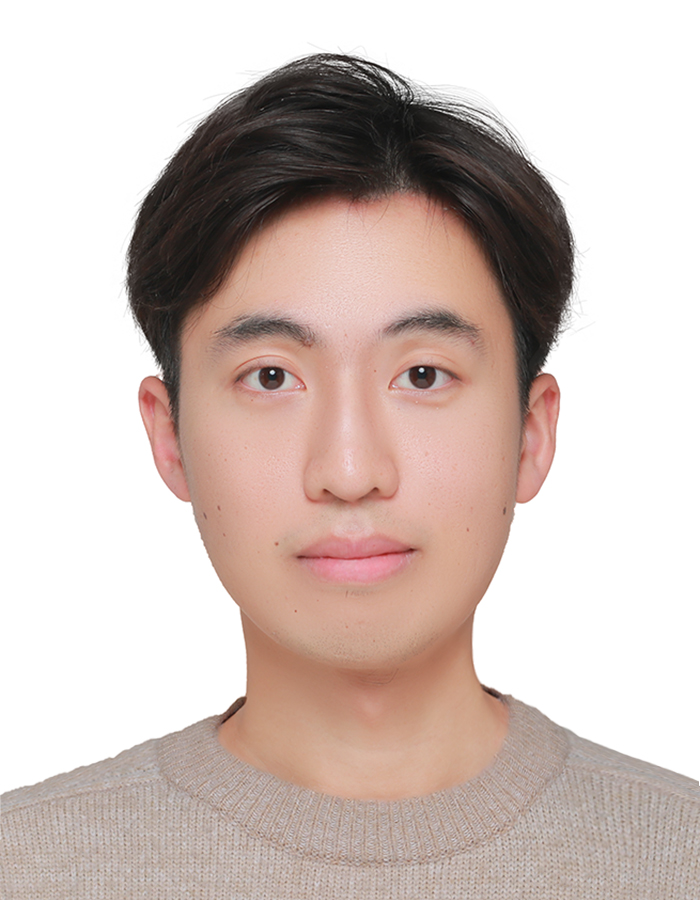}}]{Mingcheng Li}
Mingcheng Li received the B.S. degree in Computer Science and Technology from Hefei University of Technology, Heifei, China, in 2021. He is currently pursuing the Ph.D. degree at the Academy for Engineering and Technology, Fudan University, Shanghai, China. His research interests include multimodal perception, large multimodal models, and causal inference.
\end{IEEEbiography}
\vspace{-1cm}

\begin{IEEEbiography}[{\includegraphics[width=1in,height=1.25in,clip,keepaspectratio]{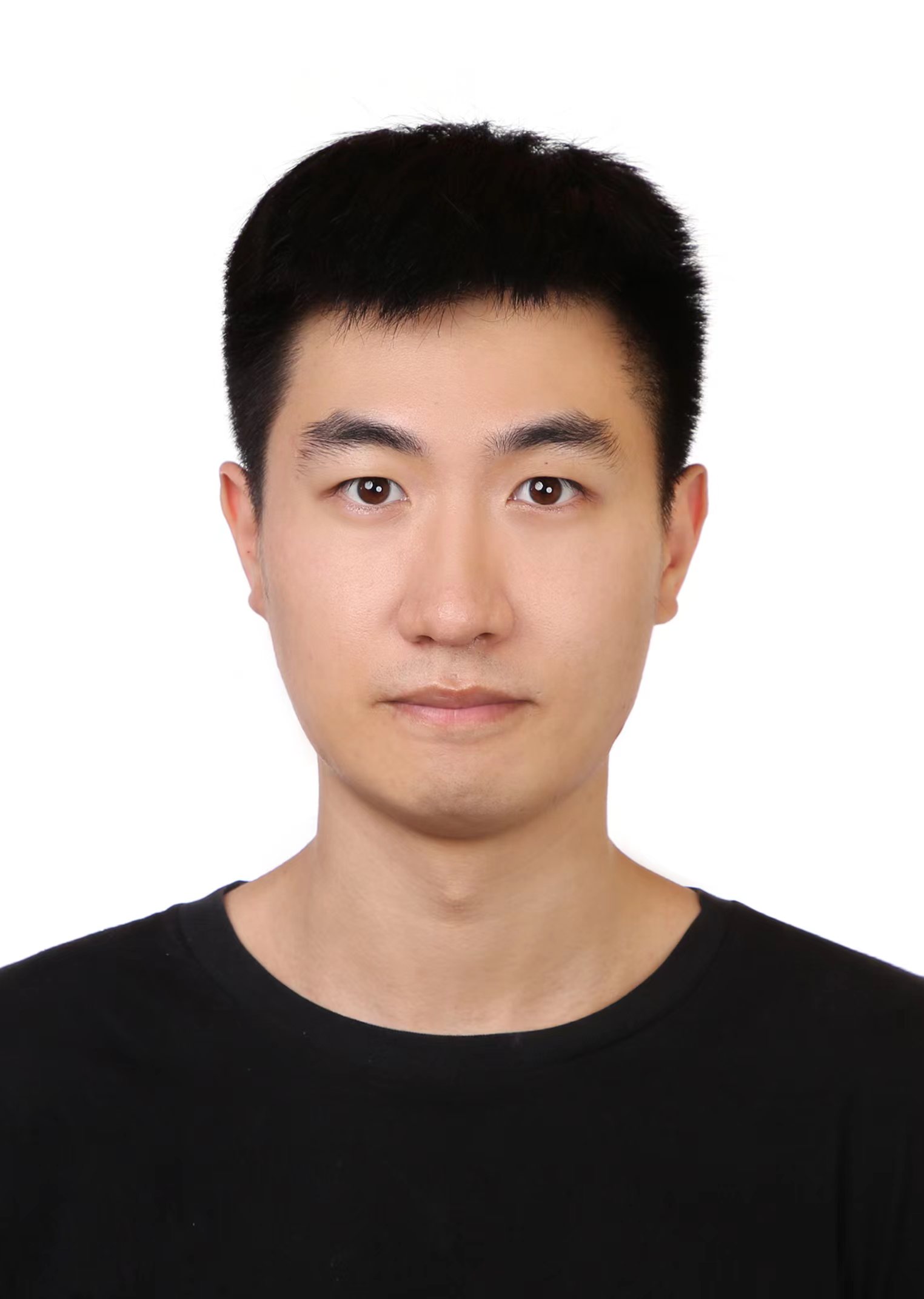}}]{Linhao Qu} is a Ph.D. student majoring in Biomedical Engineering at the School of Basic Medical Sciences, Fudan University. His research focuses on computational pathology, medical image processing, information fusion, and data mining. He has published over 20 papers in renowned international conferences and journals, appearing in prestigious international academic conferences such as NeurIPS, ICCV, CVPR, AAAI, MICCAI, and reputable journals like IEEE TMI, IEEE JBHI, Information Fusion, Modern Pathology and American Journal of Pathology.
\end{IEEEbiography}
\vspace{-1cm}

\begin{IEEEbiography}[{\includegraphics[width=1in,height=1.25in,clip,keepaspectratio]{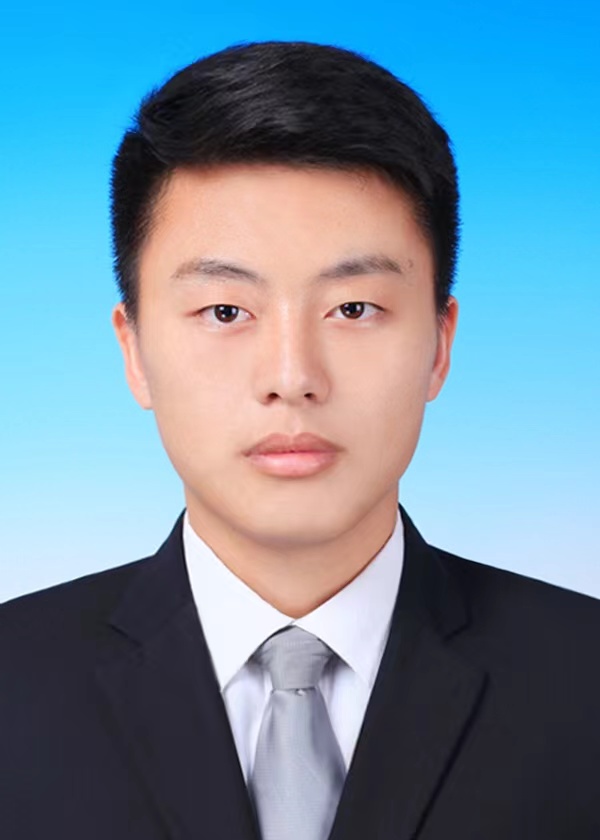}}]{Kun Yang} received the B.S. degree in Automation from the Donghua University, Shanghai, China, in 2020. He is currently pursuing the Ph.D. degree at the Academy for Engineering and Technology, Fudan University, Shanghai, China. His research interests include collaborative perception, autonomous vehicles, and vehicular edge computing. He is currently working at the Fudan Institute on Networking Systems of AI, Fudan University.
\end{IEEEbiography}
\vspace{-1cm}

\begin{IEEEbiography}[{\includegraphics[width=1in,height=1.25in,clip,keepaspectratio]{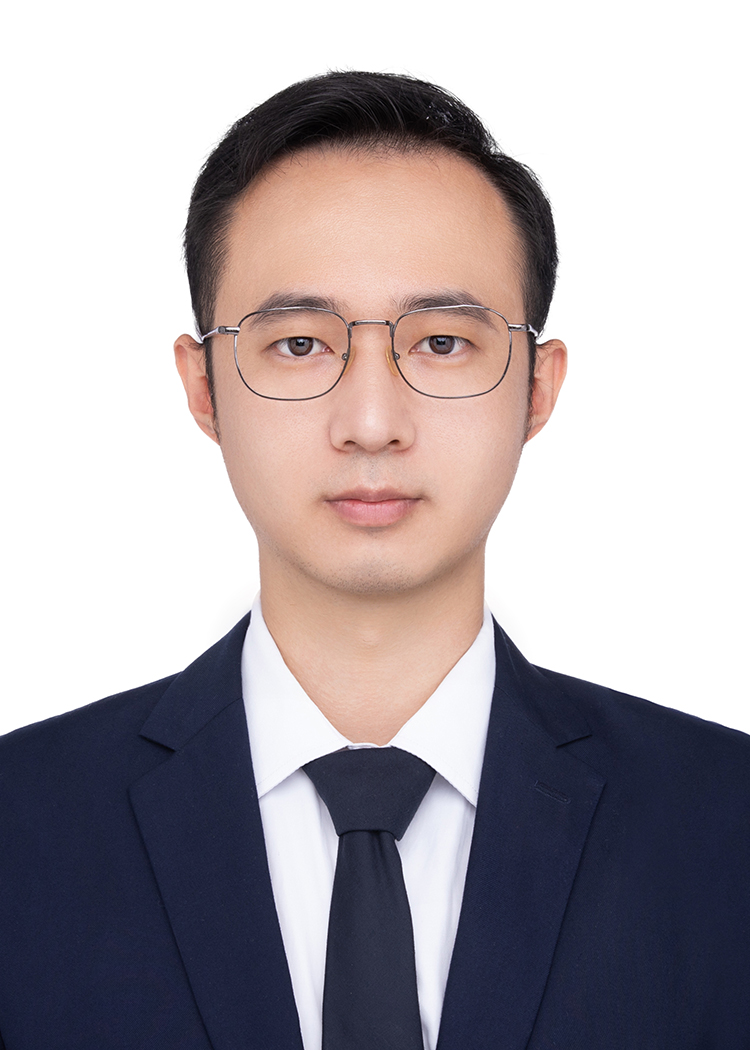}}] {Peng Zhai} received the Ph.D. degree in technology for computer applications from Fudan University, Shanghai, China, in 2022. He is currently a Postdoc at the Academy for Engineering and Technology of Fudan University. His research interests include control theory, reinforcement learning, machine intuition, and robot control.
\end{IEEEbiography}
\vspace{-1cm}

\begin{IEEEbiography}[{\includegraphics[width=1in,height=1.25in,clip,keepaspectratio]{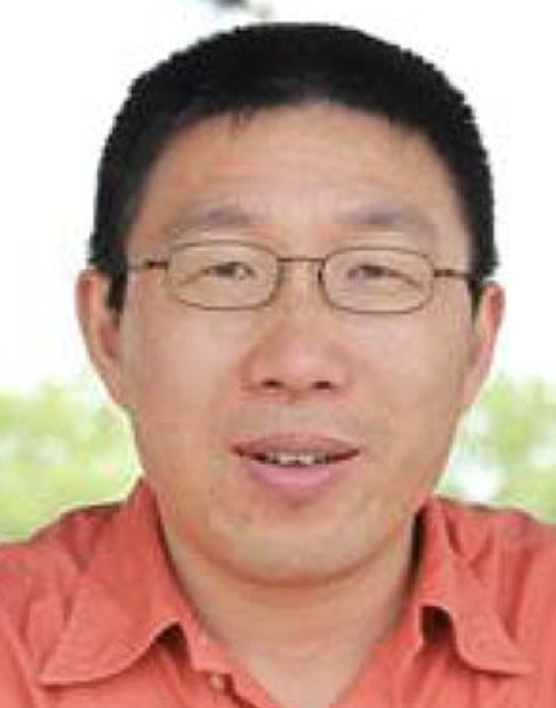}}] {Song Wang} (Senior Member, IEEE) received the Ph.D. degree in electrical and computer engineering from the University of Illinois at Urbana Champaign
(UIUC), Champaign, IL, USA, in 2002. He was
a research assistant with the Image Formation and
Processing Group, Beckman Institute, UIUC, from
1998 to 2002. In 2002, he joined the Department
of Computer Science and Engineering, University of
South Carolina, Columbia, SC, USA, where he is
currently a Professor. His current research interests include computer vision, image processing, and machine learning. He is also serving as the Publicity Chair/the Web Portal Chair
of the Technical Committee of Pattern Analysis and Machine Intelligence of the IEEE Computer Society and an associate editor for IEEE Transaction on
Pattern Analysis and Machine Intelligence, IEEE Transaction on Multimedia, and Pattern Recognition Letters.
\end{IEEEbiography}

\begin{IEEEbiography}[{\includegraphics[width=1in,height=1.25in,clip,keepaspectratio]{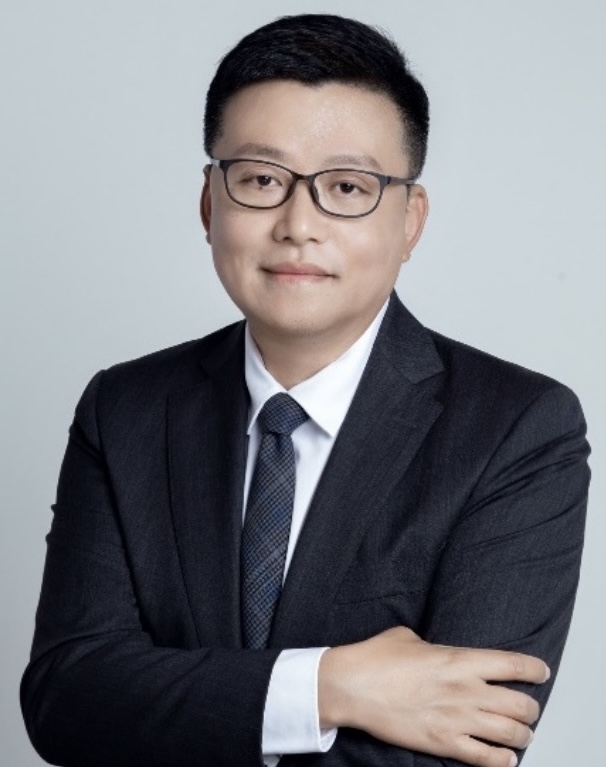}}]{Lihua Zhang} received the Ph.D. degree from the Department of Automation, Tsinghua University, Beijing, China, in 2000. He is currently a Professor at the Academy for Engineering and Technology, Fudan University, Shanghai, China. In recent years, he has participated in a number of national science and technology research and development projects as a project leader and sub-project. His current research interests are in artificial intelligence and its applications, including machine intuition, computer vision and intelligent perception, virtual reality and digital twinning, intelligent robotics and unmanned systems, intelligent computing and intelligent chips, intelligent healthcare, intelligent connected vehicles, etc.
\end{IEEEbiography}

\end{document}